\documentclass[lettersize,journal]{IEEEtran}
\usepackage{amsmath,amsfonts}
\usepackage{algorithmic}
\usepackage{algorithm}
\usepackage{array}
\usepackage[caption=false,font=normalsize,labelfont=sf,textfont=sf]{subfig}
\usepackage{textcomp}
\usepackage{stfloats}
\usepackage{url}
\usepackage{verbatim}
\usepackage{graphicx}
\usepackage{cite}
\usepackage{multirow}
\usepackage{booktabs}
\usepackage{makecell}
\usepackage{threeparttable}
\usepackage{bm}
\usepackage{color}
\usepackage{utfsym} 
\usepackage[symbol]{footmisc}
\usepackage{hyperref}
\usepackage{colortbl}

\definecolor{lightgreen}{rgb}{0.74, 0.988, 0.788}
\definecolor{lightyellow}{rgb}{1.0, 0.92157, 0.8039}
\definecolor{lightblue}{rgb}{0.6863, 0.87843, 0.90196}

\begin{document}
	
	\title{CUAHN-VIO: Content-and-Uncertainty-Aware Homography Network for Visual-Inertial Odometry}
		
	\author{Yingfu Xu and Guido C.\thinspace H.\thinspace E. de Croon,~\IEEEmembership{Member,~IEEE}
	\thanks{The authors are with the Micro Air Vehicle Laboratory, Faculty of Aerospace Engineering, Delft University of Technology, The Netherlands. (emails: \href{mailto:y.xu-6@tudelft.nl}{\texttt{y.xu-6@tudelft.nl}}; \href{mailto:G.C.H.E.deCroon@tudelft.nl}{\texttt{G.C.H.E.deCroon@tudelft.nl}}). }}
		
	\maketitle
	
	\begin{abstract}
		Learning-based visual ego-motion estimation is promising yet not ready for navigating agile mobile robots in the real world. 
		In this article, we propose CUAHN-VIO, a robust and efficient monocular visual-inertial odometry (VIO) designed for micro aerial vehicles (MAVs) equipped with a downward-facing camera.
		The vision frontend is a content-and-uncertainty-aware homography network (CUAHN) that is robust to non-homography image content
		and failure cases of network prediction. It not only predicts the homography transformation but also estimates its uncertainty.
		The training is self-supervised, so that it does not require ground truth that is often difficult to obtain. The network has good generalization that enables ``plug-and-play" deployment in new environments without fine-tuning. 
		A lightweight extended Kalman filter (EKF) serves as the VIO backend and utilizes the mean prediction and variance estimation from the network for visual measurement updates.
		CUAHN-VIO is evaluated on a high-speed public dataset and shows rivaling accuracy to state-of-the-art (SOTA) VIO approaches. Thanks to the robustness to motion blur, low network inference time ($\sim$23ms), and stable processing latency ($\sim$26ms), CUAHN-VIO successfully runs onboard an Nvidia Jetson TX2 embedded processor to navigate a fast autonomous MAV.
		
	\end{abstract}
	
	\begin{IEEEkeywords}
		Deep homography, self-supervised learning, uncertainty estimation, visual-inertial odometry, micro air vehicle.
	\end{IEEEkeywords}

	\section*{Supplementary Materials}
	 A supplementary document can be downloaded at \href{https://github.com/tudelft/CUAHN-VIO}{https://github.com/tudelft/CUAHN-VIO}. The code developed for CUAHN-VIO will be open-sourced at the same link upon the publication of this article. A video of VIO runtime performance is available at \href{https://youtu.be/_NgDkgON-nE}{https://youtu.be/\_NgDkgON-nE}.
	
	\section{Introduction} \label{intro}
	
	\begin{figure}[!hbpt]
		\centering
		\makebox{
			\includegraphics[scale=0.395]{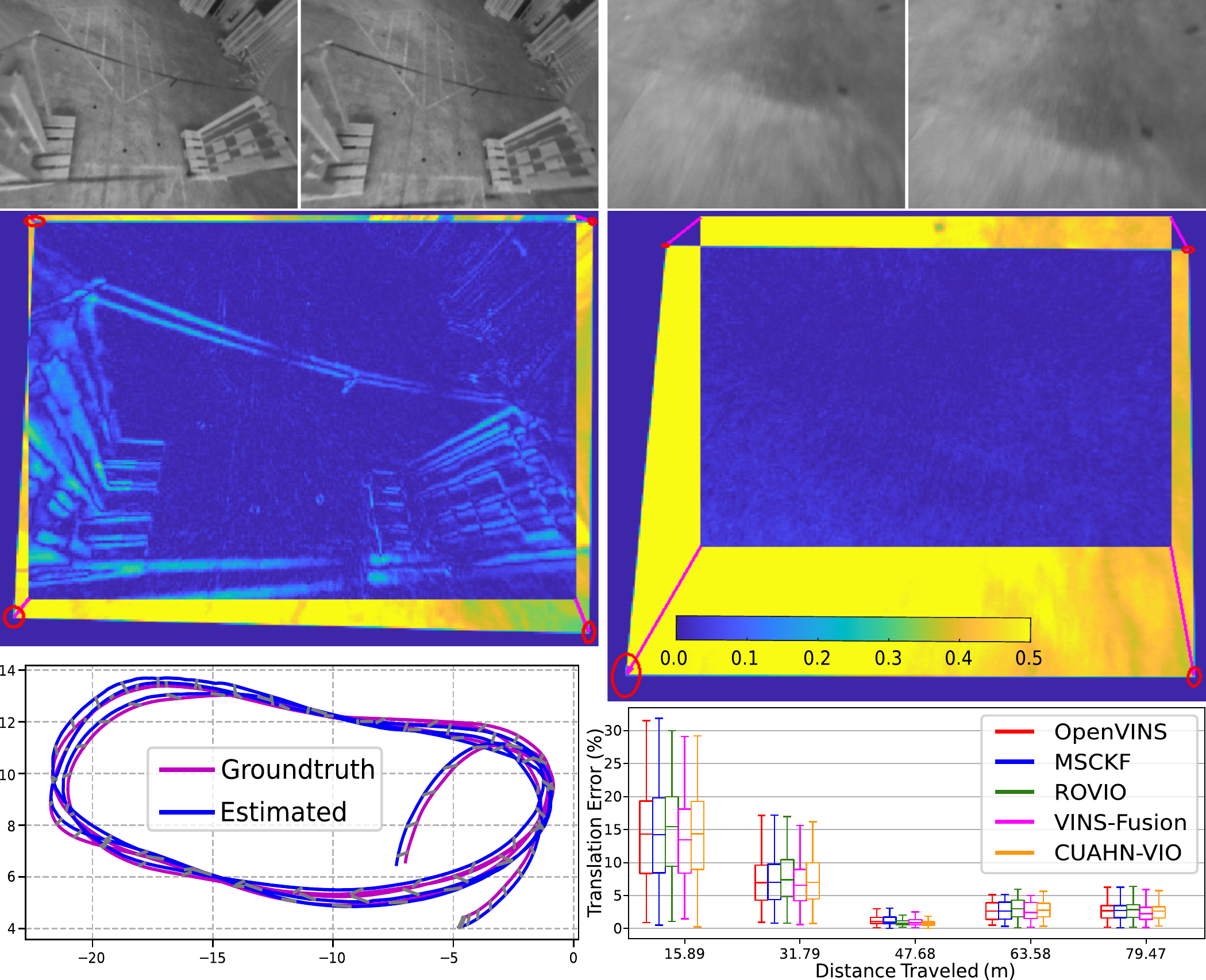}}
		\caption{Visualized outputs of CUAHN-VIO evaluated on Seq. 13 of the UZH-FPV dataset \cite{delmerico2019we}. CUAHN is robust to sparse non-planar objects and motion blur. Example image pairs are respectively shown on the left and right of the first row. The colormaps (2nd row) show the photometric error between the current image and the previous image warped according to the homography transformation predicted by the network. The arrows in pink are the network-predicted optical flow vectors of the four corner pixels. The red ellipses are the 95\% confidence ellipses of the endpoint distributions of the optical flow vectors. They are plotted according to the uncertainty estimation from the network. The boxplot in the bottom right shows the relative translation errors of different sub-trajectory lengths. CUAHN-VIO rivals SOTA approaches. 
		}
		\label{fig1}
	\end{figure}
	
	\IEEEPARstart{T}{hanks}
	to the rapid development of computer vision and state estimation techniques, VIO has become a trustworthy component of autonomous robots, such as MAVs. It expands the application scope of MAVs to GPS-denied environments such as indoor spaces. Monocular VIO is attractive to MAVs because it only requires an inertial measurement unit (IMU) and a single camera.
	
	Traditional monocular VIO is built upon projective geometry.
	Feature-based approaches \cite{campos2021orb, li2013high, leutenegger2015keyframe, qin2017vins, geneva2020openvins, qiu2020lightweight} detect and track handcrafted feature points along image frames. Direct approaches \cite{engel2014lsd, engel2017direct, zhong2020efficient} directly utilize the photometric intensities of pixels. Hybrid approaches \cite{bloesch2017iterated, forster2016svo} combine both.
	Although such approaches has been widely recognized, their vision frontends have inherent defects. 
	They are often affected by disadvantageous and hard-to-model environmental factors, such as motion blur, varying illumination, and textureless regions.

	An alternative is learning to predict camera ego-motion by a deep neural network (DNN). As observed in \cite{wang2021tartanvo, wagstaff2022self, xu2021cnn, costante2015exploring, wang2018end, clark2017vinet}, DNNs better cope with visually degraded conditions than their handcrafted counterparts.
	Often referred to as \textit{PoseNet}, the DNN regresses to the 6-dimensional (6-d) relative pose, \textit{i.e.} 3-d rotation and 3-d translation, between temporally consecutive camera views. The network input is a concatenation of images or an optical flow map, where the camera ego-motion is encoded. 
	In supervised learning, \textit{PoseNet} learns translational motion with metric scale from ground-truth labels \cite{costante2015exploring, wang2018end, parameshwara2022diffposenet, clark2017vinet, han2019deepvio, li2020towards}. But the labels are often expensive to obtain and thus limit the amount of training data.
	Alternatively, self-supervised learning can be conducted by involving co-training with another network that predicts a pixel-wise depth map \cite{zhou2017unsupervised, yin2018geonet, chen2019self, mahjourian2018unsupervised, bian2019unsupervised, li2018undeepvo, yang2020d3vo, almalioglu2022selfvio, wagstaff2022self}. The training loss derives from the difference between the actually captured image and the ``virtual" one synthesized by image warping according to the predicted relative pose and depth. 
	
	When training with monocular videos, translation and depth are scaled mutually to best explain the visual correspondences within the input images. Since there is no constraint on the scales in the loss function, as pointed out in \cite{bian2019unsupervised}, networks not only suffer from scale ambiguity but also have scale-inconsistent predictions over different video snippets.
	Metric scale can be learned from calibrated stereo images \cite{li2018undeepvo, yang2020d3vo} or videos with synchronized IMU data streams \cite{wagstaff2022self}. But both methods raise higher demands on training data.

	Besides the issue of scale discussed above, we believe that learning-based ego-motion estimation has three major challenges on the road to 
	being trusted in deployment onboard MAVs. 
	The first one is the network generalization capacity. Obviously, the requirement of fine-tuning in every new environment is a fatal barrier to wider application.
	However, to the best of our knowledge, only three works \cite{xu2021cnn, wang2021tartanvo, parameshwara2022diffposenet} demonstrated cross-dataset generalization. All of them utilized large datasets synthesized in simulation.
	In most works, the networks are trained and tested on the same dataset. The most popular is KITTI \cite{geiger2012we}, a car dataset with three degrees-of-freedom (3-DoF) motion. When it comes to a smaller number of training samples and more difficult motion patterns, networks \cite{wang2018end, han2019deepvio, clark2017vinet, li2020towards, wagstaff2022self} show worse accuracy than traditional approaches on EuRoC \cite{burri2016euroc}, an indoor MAV flight dataset with 6-DoF motion.
	
	The second challenge is network prediction uncertainty. It is typical that most deep learning application works purely pursue prediction accuracy on certain datasets. 
	It is not enough because we lack knowledge of the mechanisms of DNNs and thus highly inaccurate predictions may appear, especially when the input sample is outside the training distribution or distorted by noise. Such outliers can cause a big drift in ego-motion estimation and mislead the robot. Uncertainty estimation can remedy this problem. For example, estimating the uncertainties of each network prediction and using them within the bundle adjustment (BA) backend lead to better accuracy than constant hand-tuned uncertainty \cite{wang2018end, yang2020d3vo}.

	The last challenge is high computation time. 
	The causes are, for instance, the network being too deep \cite{wang2018end}, combining multiple networks that together are very large \cite{almalioglu2022selfvio}, or using an expensive intermediate representation such as a dense optical flow map \cite{costante2015exploring, wang2021tartanvo}.
	Works \cite{costante2015exploring, wang2018end, almalioglu2022selfvio, wang2021tartanvo} reported network inference time of more than 40ms measured on Nvidia GPUs designed for desktop computers.

	In this article, we propose CUAHN-VIO that overcomes the three challenges to a large extent. 
	Instead of \textit{PoseNet}, the vision frontend is a network that predicts the planar homography transformation. It is a pixel-level task and thus generalizes better across cameras with different intrinsics.
	The network input is a pair of temporally consecutive images captured by a downward-facing camera mounted on an MAV. 
	We show cross-dataset evaluation and real-world flight experiments without any fine-tuning to demonstrate the decent generalization of the network.
	The network prediction uncertainty is estimated with minor extra computation. It strengthens the system's robustness toward outlier predictions and contributes significantly to VIO accuracy.
	In terms of inference time, CUAHN-VIO runs faster than 30 frame-per-second (fps) onboard an Nvidia Jetson TX2 mobile processor. Its robustness toward high-speed motion is highlighted in a comparative experiment with a traditional VIO approach.

	The main contributions of this work can be summarized as:
	\begin{itemize}
		\item We propose a practical scheme of training a homography network in a self-supervised fashion. It has high prediction accuracy, high-quality uncertainty estimation, and robustness toward sparse 3-d structures in view.
		
		\item We build a VIO system upon the network and an EKF-based backend. The metric scale is maintained by the integration of IMU measurements. The network architecture of cascaded blocks makes full use of the EKF \textit{a priori} state, contributing to both accuracy and efficiency.
		
		\item To the best of our knowledge, CUAHN-VIO is the first learning-based VIO that not only rivals SOTA approaches in both accuracy and efficiency but also has ``plug-and-play" generality and convenience for robot navigation in the real world.
	\end{itemize}

	\section{Related Works}
	
	\subsection{Learning-based Visual Ego-Motion Estimation} \label{literature_egomotion}
	
	For \textit{PoseNets} learning to predict 3-d translational motion in metric scale from monocular video \cite{costante2015exploring, wang2018end, parameshwara2022diffposenet, li2018undeepvo, yang2020d3vo},
	the scale in testing is recovered by the network's ``memory" of the scene structure, \textit{e.g.} the size of objects, in the training set.  
	When testing in a new environment,
	the scale is possibly inaccurate because of the non-perfect generalization. An extreme case is a miniature park. A car model may be misidentified by the network as a real car that the network has seen in training. Consequently, a translation of a few centimeters may be mistaken for meters of motion.
	To avoid this problem, TartanVO \cite{wang2021tartanvo} recovers up-to-scale translational motion by constraining the normalized translation vector in training.
	The network is trained on a large-scale dataset of simulation environments and generalizes well to real-world datasets.
	A potential problem is that the normalized translation is indefinite when the camera is close to stationary or in pure rotation. In the evaluation of \cite{wang2021tartanvo}, the scale of predicted translation is recovered by ground truth metric scale. So the potential bad effect cannot be observed. Another drawback of this approach is that calculating the optical flow map that is the input of \textit{PoseNet} is computationally heavy.
	
	When IMU measurements are fed along with video, two separate subnetworks can be respectively in charge of visual and inertial processing at different sensor rates and output two intermediate tensors. 
	And another subnetwork takes the concatenated intermediate tensors as input to perform sensor fusion and pose prediction \cite{clark2017vinet, han2019deepvio}. 
	This setup has a principle-level generalization issue.
	Networks for IMU processing and sensor fusion implicitly ``remember" the sensor setup of the training set, \textit{e.g.} bias and noise characteristics of IMU and the extrinsics between IMU and camera. Generalizing to a new sensor setup is difficult. 
	IMU data is low-dimensional and has well-understood models that are grounded in physics.
	Practising this idea, an end-to-end supervised learning scheme for a loosely-coupled VIO is proposed in \cite{li2020towards}. 
	Its backend is a differentiable EKF whose states are propagated by integrating IMU measurements.

	For self-supervised learning, SfMLearner \cite{zhou2017unsupervised} firstly proposed to simultaneously train two networks that respectively predict ${T}_{t \rightarrow s}$ and $\mathcal{D}_t$. ${T}_{t \rightarrow s}$ is the relative pose between source image $\mathcal{I}_s$ and target image $\mathcal{I}_t$. $\mathcal{D}_t$ is the pixel-wise depth map of $\mathcal{I}_t$.
	An image $\widetilde{\mathcal{I}}_{s}$ can be synthesized by warping $\mathcal{I}_s$ according to the 
	2-d projections of the 3-d point cloud established from $\mathcal{D}_t$ on the image plane of $\mathcal{I}_s$ located at ${T}_{t \rightarrow s}$.
	Based on the assumption that the pixels in consecutive images corresponding to the same point in the scene have the same intensity, 
	the supervision signal derives from the photometric difference between $\widetilde{\mathcal{I}}_{s}$ and $\mathcal{I}_t$. We call it reprojection-based loss for simplicity.
	This scheme was further developed by also predicting $\mathcal{D}_s$ and punishing the 3-d geometric inconsistency between $\mathcal{D}_t$ and $\mathcal{D}_s$ \cite{mahjourian2018unsupervised, bian2019unsupervised}.
	
	Also using reprojection-based loss, SelfVIO \cite{almalioglu2022selfvio} performs self-supervised learning of a depth network and three subnetworks for pose prediction. They have the same functions as the subnetworks of supervised-learning VIO \cite{clark2017vinet} and \cite{han2019deepvio}.
	IMU measurements bring in motion information with metric scale, however, as pointed out in \cite{wagstaff2022self}, the IMU processing network has no knowledge of the physical model of IMU and the reprojection-based loss does not account for scale.
	So the metric scale of the IMU measurements is transformed by the trained network and thus predictions still have no metric scale. 
	Extended from \cite{li2020towards},
	the \textit{PoseNet} prediction in \cite{wagstaff2022self} is also fused with the IMU-propagated \textit{a priori} states by an EKF. The refined \textit{a posteriori} ego-motion and the output of a depth network together minimize the self-supervised reprojection-based loss in training. 
	The metric scale is obtained by explicitly integrating IMU measurement according to its physical model.
	But the authors used a 7-DoF similarity transformation (\textit{Sim3}) for trajectory alignment to quantify the VIO accuracy. So we do not know how well the scale of their VIO output matches the metric scale.
	
	\subsection{Network Uncertainty Estimation in Computer Vision} \label{literature_uncertainty}
	
	According to the taxonomy of \cite{kendall2017uncertainties}, for a deep network model, there are two major types of uncertainty that can be modeled.
	\textit{Aleatoric} uncertainty captures noise inherent in the network input.
	It can be learned from the real data distribution by the network \cite{nix1994estimating}. The loss function is the negative log-likelihood (NLL) loss. We referred to it as predictive uncertainty to emphasize the way by which it is obtained.
	
	\textit{Epistemic} uncertainty reflects the ignorance about the \textit{perfect} model that maps clean noiseless input to the desired output. It can be explained away given enough data. 
	Bayesian neural networks \cite{neal1995bayesian} model the trainable network parameters as distributions instead of deterministic values to explain the \textit{Epistemic} uncertainty in the parameters.
	Since exact Bayesian inference is computationally intractable for DNNs \cite{lakshminarayanan2017simple, maddox2019simple}, practical strategies of approximate inference were developed such as ensembles of DNNs (deep ensembles) \cite{lakshminarayanan2017simple} and Monte Carlo Dropout (MC-Dropout) \cite{gal2016dropout}. 
	Given a certain input, these methods estimate the distribution of the network prediction by combining the multiple outputs of an empirically sampled subset of all the possible network instances. 
	
	The models of deep ensembles are different point estimates (instead of distribution) of model parameters. They are trained independently to de-correlate their predictions.
	Ensemble members can be trained on different randomly sampled subsets of the entire training set, referred to as bootstrapping. 
	By contrast, MC-Dropout requires only a single network model trained with dropout. Also deploying dropout in inference, multiple independent models are randomly sampled via multiple forward passes.
	\textit{Epistemic} uncertainty is referred to as empirical uncertainty in this article to highlight its acquisition approach.
	
	The above introduced uncertainty estimation approaches have been applied to computer vision tasks.
	Predictive uncertainty has been proven effective in the prediction of object pose \cite{kaufmann2019beauty}, camera ego-motion \cite{klodt2018supervising, peretroukhin2019deep, li2020towards, wagstaff2022self, wang2018end}, monocular depth \cite{poggi2020uncertainty, klodt2018supervising, kendall2017uncertainties}, optical flow \cite{ilg2018uncertainty}, semantic segmentation\cite{kendall2017uncertainties}, and image classification \cite{lakshminarayanan2017simple}.
	For empirical uncertainty, deep ensembles were evaluated in image classification \cite{lakshminarayanan2017simple}, optical flow \cite{ilg2018uncertainty}, and monocular depth \cite{poggi2020uncertainty}. 
	Likewise, MC-Dropout was adopted in networks for optical flow \cite{ilg2018uncertainty}, monocular depth \cite{kendall2017uncertainties, poggi2020uncertainty}, semantic segmentation \cite{kendall2017uncertainties}, and camera pose regression \cite{kendall2016modelling}.

	Our purpose in studying network uncertainty estimation is for a better knowledge of visual measurement to benefit Bayesian state estimation. With the similar aim, 
	Kaufmann \textit{et al.}  \cite{kaufmann2019beauty} fuse the network-predicted gate pose and its uncertainty with outputs of a VIO system by an EKF.
	The purpose is to compensate for the gate displacement and the accumulating error of VIO in autonomous drone racing.  
	Embedded in a traditional VO, D3VO \cite{yang2020d3vo} leverages the predictive uncertainty. The uncertainty map of photometric matching acts as the weights of the photometric energy in the BA backend. The relative pose network of D3VO has no uncertainty estimation, so the weights in the optimization of pose energy are set as constant. 
	In \cite{wang2018end}, six predictive standard deviations of 6-DoF relative pose are used in BA that optimizes a pose graph and achieve higher accuracy than constant hand-tuned covariance.
	Differently, Li \textit{et al.} \cite{li2020towards} proposed to learn predictive uncertainty through the Bayesian nature of a differentiable EKF instead of the widely used NLL loss. The supervision signal is the gradient flow coming from the \textit{a posteriori} ego-motion that is a function of the measurement noise covariance matrix $\bm{R}$ in EKF updating.
	Since the \textit{a posteriori} states are functions of the whole filter, the learning of $\bm{R}$ is implicitly affected by the EKF hyperparameters, \textit{e.g.} the process noise covariance matrix $\bm{Q}$, which poses a potential of overfitting.
	
	Most works adopt uncertainty estimation in supervised learning.
	About self-supervised learning, Poggi \textit{et al.} \cite{poggi2020uncertainty} made a step in the field of monocular depth. A strategy called Self-Teaching was proposed to decouple depth from pose.
	The network that outputs predictive uncertainty is trained by the NLL loss and supervised by the outputs of an already trained depth network with the same architecture.
	{The self-supervised EKF-based VIO \cite{wagstaff2022self} learns predictive uncertainty of relative camera pose from the error of \textit{a posteriori} ego-motion, same as the supervised VIO \cite{li2020towards}.
	Because the current network prediction affects the later \textit{a posteriori} states, the network is supposed to adjust the current covariance prediction according to the error of \textit{a posteriori} ego-motion in the future. So sequential training data having enough length is required.}
	
	\subsection{Deep Planar Homography} 
	
    When a camera films a 3-d point on a planar surface from different poses, the 2-d projections of this point on the image planes can be mapped by a planar homography transformation. It is a function of the ego-motion of the camera and thus useful for a VIO system.
	It can be inferred by a DNN from an input image pair. Both supervised \cite{detone2016deep, erlik2017homography, le2020deep} and self-supervised \cite{nguyen2018unsupervised, zhang2020content} learning schemes have been proposed.
	
	A planar homography transformation can be based on visual correspondences between the image pair. Multiple cascaded network blocks can predict the transformation parameters incrementally \cite{erlik2017homography,le2020deep, xu2021cnn}. 
	In this scheme, image warping and synthesizing operation is inserted between every two adjacent blocks. After the inference of each block, an image is synthesized by warping the original one using bilinear interpolation \cite{jaderberg2015spatial} according to the prediction(s) of the previous block(s).
	The next block infers from the synthesized image and the other image. Between them, there are supposed to be fewer visual disparities than the original image pair. 
	In this way, each block predicts a part of the total transformation.
	Compared with a single deep network, this strategy can lead to higher accuracy and less difficulty in training thanks to the involvement of geometric knowledge and shallower architectures of network blocks.

	In many applications, it is not the case that all the visual correspondences can be explained by homography transformation. Masking out the non-homography pixels, \textit{e.g.} the ones filming 3-d structures or dynamic objects, has the goal of boosting accuracy.
	In \cite{le2020deep}, a convolutional decoder is added to the homography network for mask prediction.
	Two masks for the input image pair are predicted together and then concatenated with the images. The concatenation is the input to the next cascaded network block. In this work, mask prediction is learned from the ground truth labels. 
	{Instead, Zhang \textit{et al.} \cite{zhang2020content} implicitly learn the mask in a self-supervised way. An extra subnetwork predicts a mask for each input image. And then the mask is multiplied with the feature map of the image. The idea behind this is that the mask can weigh down the influence of non-homography pixels. The homography network infers from the mask-weighted feature maps for the transformation that is constrained by the self-supervised loss function.}
	
	\begin{figure}[b]
		\centering
		\makebox{
			\includegraphics[scale=0.67]{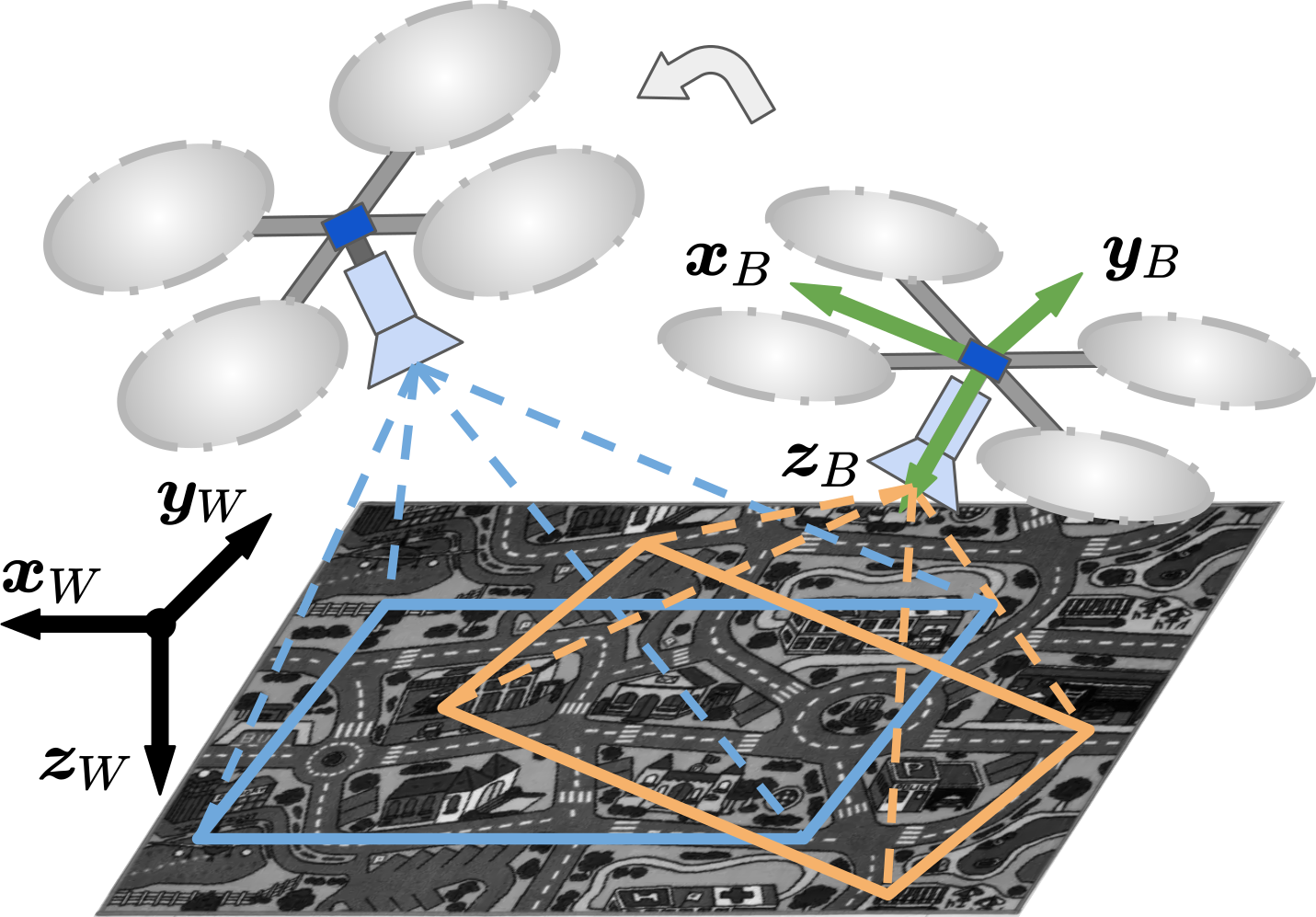}}
		\caption{An overview illustration of the application scenario, sensor setup, and coordinate definition. \textsl{W} stands for world frame and \textsl{B} stands for body frame, \textit{i.e.}, IMU frame.}
		\label{mav_cam}
	\end{figure}
	
	\section{System Overview} \label{overview}

\begin{figure}[!hbpt] 
		\centering
		\makebox{
			\includegraphics[scale=0.58]{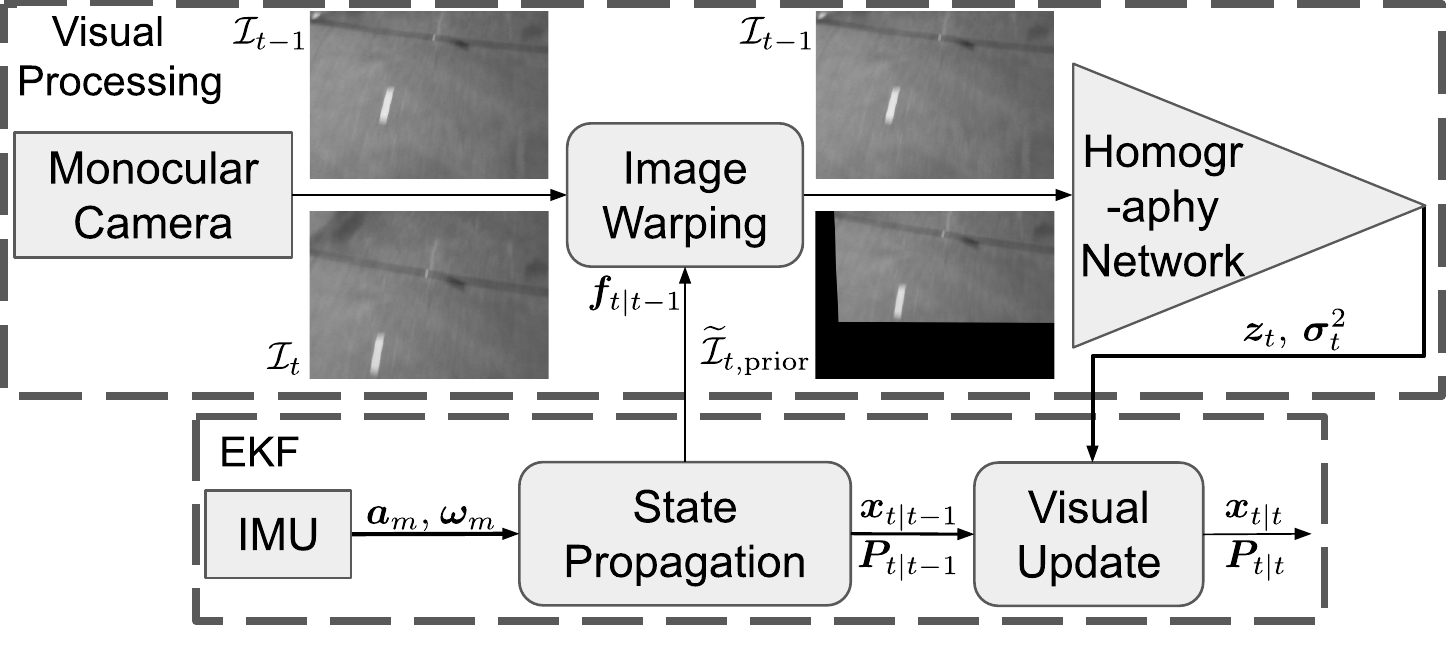}}
		\caption{An overview data flow diagram of CUAHN-VIO.}
		\label{system_overview}
	\end{figure}

	Fig. \ref{mav_cam} illustrates that CUAHN-VIO applies to MAVs that are equipped with an IMU and a downward-facing monocular camera. 
	From a pair of temporally consecutive images, the vision frontend, a DNN (Fig. \ref{arch_fig}), predicts the planar homography transformation and the uncertainty. They are utilized in updating the EKF backend as shown in Fig. \ref{system_overview}.

	Learning-based VIO approaches \cite{clark2017vinet, han2019deepvio, almalioglu2022selfvio, li2020towards, wagstaff2022self} perform end-to-end learning, \textit{i.e.}, the ego-motion inferred from both inertial and visual measurements is under constraint in the loss function. The whole VIO system is obtained from a single training attempt. But there are disadvantages. First, videos with synchronized IMU streams are required in training. They are expensive to collect, which places a barrier to enlarging the training set. Besides, extra work is required to obtain the initial poses of data sequences in self-supervised learning \cite{wagstaff2022self}.
	Second, although training the VIO submodules together contributes to in-domain accuracy, the VIO system can overfit the sensor setup of the training set.
	
	By contrast, the DNN of CUAHN-VIO is trained alone, totally decoupled from the VIO system.
	The benefit is the better generalization capacity.
	The network has no requirement for camera intrinsics or the camera-IMU extrinsics. Changes to the sensor set only require modifying the backend parameters without any change in the network.
	Besides, we do not require sequential training data. A large number of easy-to-obtain simulation image pairs (Subsection \ref{dataset}) enable the network to generalize to real-world scenes without any fine-tuning. 

	In the context of no ground-truth label, our approach requires training two networks. They are the student network acting as the VIO frontend and the teacher network.  The teacher network has more layers than the student network to gain more accuracy. It is trained by a self-supervised loss function based on photometric matching (Subsection \ref{cascaded_blocks}). Content-aware pixel-wise masks are predicted to mitigate the negative impacts of the pixels whose photometric error cannot be reduced by a better homography transformation (Subsection \ref{mask}). The teacher network is required because its mean value predictions of homography transformations are needed by the student network as targets to learn predictive uncertainty by the NLL loss (Subsection \ref{pred_uncertainty}). The student network estimates empirical uncertainty by deep ensembles or MC-Dropout (Subsection \ref{empirical_uncertainty}). Uncertainty estimation turns out to be important in improving the VIO accuracy.

	The backend of CUAHN-VIO is a simple extended Kalman filter (EKF), as shown in Fig. \ref{system_overview} and introduced in detail in Subsection \ref{ekf_backend}. It is propagated by IMU integration that explicitly maintains the metric scale. 
	The network-predicted homography transformation $\bm{z}_t$ and its uncertainty $\bm{\sigma}^2_t$ update the filter at the frame rate. 
	The \textit{a priori} homography transformation parameterized as four optical flow vectors ${\bm f}_{t|t-1}$ is utilized for pre-warping the current image $\mathcal{I}_{t}$. 
	The new image $\widetilde{\mathcal{I}}_{t,\text{prior}}$ synthesized by warping is more similar to the previous image $\mathcal{I}_{t-1}$ unless the EKF totally diverges. The smaller visual disparities make the task of the network easier. {This is especially helpful in fast flight when the optical flow is big.} With this prior information, running fewer network blocks produces higher accuracy.
	
	\section{Planar Homography Network} \label{homo_network}
	
	\subsection{Datasets} \label{dataset}

	The training dataset is the same as \cite{xu2021cnn}. 
	It is a big-scale (more than 80 thousand training samples) synthetic dataset with a wide variety of textures, realistic motion blur, and diverse motion patterns. It consists of independent image pairs with small baselines filming perfectly planar surfaces.
	We refer to it as the Basic Dataset in this article. 

	To involve non-planar and dynamic content, we collected a flight dataset by a MYNT EYE D1000-120 camera downward-facing mounted on a quadrotor MAV. It has 20 videos in which 44,837 image pairs were selected for training, 3,904 for validation, and 4,577 for testing. We put many objects of various heights on the floor that the camera filmed. Some of them moved due to the downwash from the MAV propellers. The ground-truth homography transformations were calculated from the camera poses measured by an OptiTrack motion capture system.
	Example images are shown in the left three columns of Fig. \ref{fig_mask}. This dataset is called the MYNT Dataset.
	
	The inputs of all networks in this article are required to be undistorted grayscale images with the resolution of 320$\times$224. There is no requirement on camera intrinsics.
	
	\subsection{Self-Supervised Cascaded Network Blocks} \label{cascaded_blocks}

\begin{figure}[t]
		\centering
		\makebox{
			\includegraphics[scale=0.92]{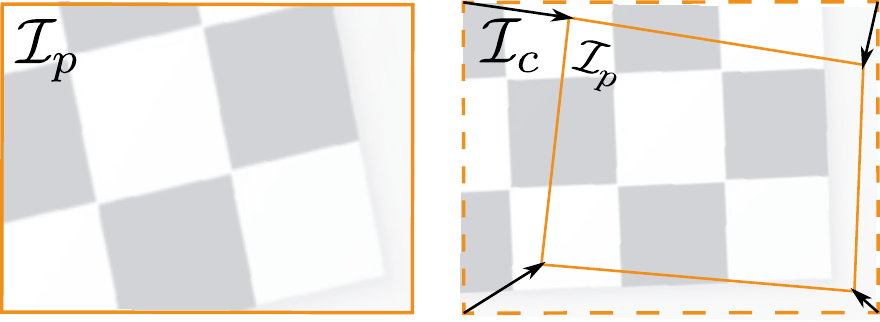}}
		\caption{An example of 8-d corner flow $\bm{f}$. Images are adapted from \cite{zhong2020direct}.}
		\label{corner_flow}
	\end{figure}

 	When the network acts as a VIO vision frontend, its input images are temporally consecutive, so we refer to them as the previous image $\mathcal{I}_p$ and the current image $\mathcal{I}_c$. 
	{The homography transformation is parameterized as four 2-d optical flow vectors in the image plane of $\mathcal{I}_c$. They point from the image corners to the pixels corresponding to the corners of $\mathcal{I}_p$, as illustrated in Fig. \ref{corner_flow}.} 
	For simplicity, they are called 8-d corner flow $\bm{f}$.
	
	\begin{figure*}[!hbpt]
		\centering
		\makebox{
			\includegraphics[scale=0.415]{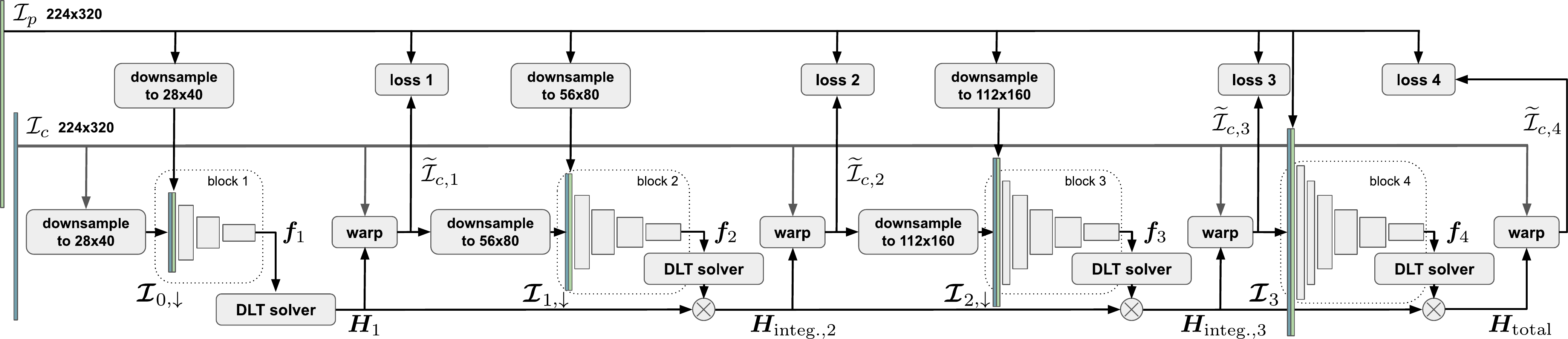}}
		\caption{The architecture of cascaded network blocks for planar homography transformation prediction. {The downward arrow in the foot marker of $\bm{\mathcal{I}}_{i,\downarrow}$ indicates that it has been downsampled. Data flows correspond to training. In inference, loss terms are not calculated and there is no DLT solver for $\bm{f}_{4}$. The network output $\bm{f}_{\text{total}}$ is obtained by \eqref{linear_transfor_f_vec}.} }
		\label{arch_fig}
	\end{figure*}
	
	As shown in Fig. \ref{arch_fig}, the proposed network has four cascaded blocks that are gradual in terms of the number of layers and the resolution of the input. The 1st block is the shallowest and its input is the most downsampled. The 4th block is the deepest and it infers from full-resolution images.
	An input tensor to a network block is made of two (downsampled) images concatenated along the channel dimension.
	The 1st block infers from the downsampled $\mathcal{I}_p$ and $\mathcal{I}_c$ and regresses to $\bm{f}_1$. The direct linear transformation (DLT\footnotemark{}) solver calculates the homography matrix $\bm{H}_1$ from $\bm{f}_1$. 
	\footnotetext{An asterisk indicates that further elaboration is available in the supplementary document. Applying to the whole article.}
	The correspondence between the float pixel coordinate $(u_c, v_c)$ in $\mathcal{I}_c$ and the integer pixel coordinate $(u_p, v_p)$ in $\mathcal{I}_p$ is 
	\begin{equation}{\label{H_transfer}}
		\lambda[u_c, v_c, 1] = \bm{H}_1 [u_p, v_p, 1]^T.
	\end{equation}
	With $(u_c, v_c)$, a new image can be synthesized by warping $\mathcal{I}_c$ using differentiable bilinear interpolation \cite{jaderberg2015spatial}. The synthesized image is referred to as $\widetilde{\mathcal{I}}_{c, 1}$, the warped $\mathcal{I}_c$ according to $\bm{f}_1$. $\widetilde{\mathcal{I}}_{c, 1}$ is then downsampled and concatenated with the downsampled $\mathcal{I}_p$ to form up the input tensor of the 2nd block. $\bm{f}_2$, the prediction of the 2nd block, is supposed to point from the corners to the pixels in $\widetilde{\mathcal{I}}_{c, 1}$ that correspond to the corners of $\mathcal{I}_p$. $\bm{H}_{2}$ is integrated with $\bm{H}_1$ by matrix multiplication to produce the updated $\bm{H}_{\text{integ.},2}$. The same processes repeat for the 3rd and 4th blocks.
	The later warping is based on the refined $\bm{H}_{\text{integ.},i}$. Thus there should be fewer discrepancies between the pair of (downsampled) images that are input to the next block. In this way, blocks running earlier are trained to capture bigger disparities and the later blocks are good at refining $\bm{H}_{\text{integ.},i}$ by inferring from the more-and-more similar images. 
	The final prediction $\bm{f}_{\text{total}}$ is the total corner flow between $\mathcal{I}_c$ and $\mathcal{I}_p$.
	It is obtained by
	\begin{equation}{\label{linear_transfor_f_vec}}
		\begin{aligned}
			\begin{split}
				\lambda (\bm{f}_{\text{total}, j} + \bm{c}_j) =  \bm{H}_{\text{integ.},3}  (\bm{f}_{4,j} + \bm{c}_j),
			\end{split}
		\end{aligned}
	\end{equation} 
	{where $j$ indexes over the four corners and $\bm{c}_j$ is the corner pixel coordinate. $(\bm{f}_{4,j} + \bm{c}_j)$ is the predicted coordinate of the pixel in $\widetilde{\mathcal{I}}_{c,3}$ corresponding to the $j$th corner of $\mathcal{I}_p$. $(\bm{f}_{\text{total}, j} + \bm{c}_j)$ is the coordinate of the pixels in ${\mathcal{I}}_{c}$ that has the same intensity as the pixel at $(\bm{f}_{4,j} + \bm{c}_j)$ in $\widetilde{\mathcal{I}}_{c,3}$.
	Here pixel coordinates are 3-d homogeneous coordinates.}
	
	The similarity between  $\widetilde{\mathcal{I}}_{c,i}$ and $\mathcal{I}_p$ indicates how accurate is $\bm{H}_{\text{integ.},i}$. The self-supervised loss function is established as \eqref{self_loss} and \eqref{pixel_loss}.
	\begin{equation}{\label{self_loss}} 
		\mathcal{L}_{i} = \frac{1}{|V|}\sum_{k\in V} \mathcal{L}(\bm{\mathcal{I}}_{i,k})
	\end{equation}
	\begin{equation}{\label{pixel_loss}}
		\mathcal{L}(\bm{\mathcal{I}}_{i,k}) = \frac{\alpha}{2}(1-{\rm SSIM}(\bm{\mathcal{I}}_{i,k})) + (1 - \alpha) \cdot |\mathcal{I}_{p,k} - \widetilde{\mathcal{I}}_{c,i,k}| 
	\end{equation}
	$V$ denotes the set of all valid pixels excluding the ones sampled outside the border of $\mathcal{I}_c$. It is a mask calculated from $\bm{H}_{\text{integ.},i}$. 
	$\bm{\mathcal{I}}_{i}$ denotes the aggregation of $\mathcal{I}_{p}$ and $\widetilde{\mathcal{I}}_{c,i}$.
	Same as other works \cite{godard2019digging, poggi2020uncertainty, klodt2018supervising, yang2020d3vo}, we involve both the $L1$ loss of pixel-wise photometric error and Structured Similarity Index Measure (SSIM) loss in the loss function of self-supervised learning, as shown in \eqref{pixel_loss} where $\alpha=0.85$. 
	Multi-stage losses are calculated from each $\bm{\mathcal{I}}_{i}$. 
	Their weights are respectively 0.1, 0.2, 0.3, and 0.4 from earlier to later blocks. 
	
	The planar homography network is implemented\footnotemark[\value{footnote}] in a Python environment with PyTorch library and trained on the Basic Dataset by the self-supervised loss. We employed bidirectional training, \textit{i.e.} concatenating an image pair in two opposite orders.
	{The average error of the predicted $\bm{f}_{\text{total}}$ on the testing set of the Basic Dataset is 0.275 pixels.}
	It is the average of the absolute values of elements of the 8-d error vector that represents the difference between the network prediction and ground truth, \textit{i.e.} $\frac{1}{8}\sum_{j=1}^{4} |\bm{f}_{j,u}-\bm{f}_{j,u,\text{GT}}| + |\bm{f}_{j,v}-\bm{f}_{j,v,\text{GT}}| $. Note that this is different from the optical flow endpoint error (EPE) utilized by other works \cite{detone2016deep, zhang2020content, le2020deep}, which is the average $L$2 distance, \textit{i.e.} $\frac{1}{4}\sum_{j=1}^{4} ||\bm{f}_{j}-\bm{f}_{j,\text{GT}}||_2$.
	The reason for element-wise averaging is that, as introduced later, the uncertainty of each element of $\bm f$ is estimated independently. The error-variance data pairs for evaluating uncertainty estimation are element-wise. 
	We refer to the trained network as the Basic Model. 
	Its average inference time cost of a single image pair is 28.20 ms in Python environment and 21.16 ms in C++\footnotemark[\value{footnote}], measured on a TX2 processor in Max-P ARM power mode. 
	
	\subsection{Content-Aware Learning} \label{mask}
	
	The major assumptions made in the self-supervised loss function \eqref{pixel_loss} are that 1) the camera is facing a single perfectly planar surface, and 2) the overlapping content of both images meets the brightness consistency constraint.
	However, these assumptions can be easily violated in the real world by 3-d structures, moving objects, occlusions, and reflective materials. 
	A straightforward idea is to learn a content-aware (CA) mask to down-weight the losses of pixels violating the assumptions. The mask is supposed to be learned without ground truth. It only acts on the loss function and thus is not required during testing.
	
	To predict such a mask, 4th block is expanded to a UNet\cite{ronneberger2015u}-like architecture with skip connections. Its convolutional layers serve as the encoder part. The upsampling decoder part is added and connected to the last convolutional layer. A single mask is inferred from $\bm{\mathcal{I}}_{3}$. {The mask-involved loss function is applied to the final homography transformation prediction $\bm{H}_{\text{total}}$.}
	The rest blocks keep their original architectures and $\bm{H}_{\text{integ.},i},\,i\in \{1,2,3\}$ are still constrained by \eqref{pixel_loss} without taking the mask into account, assuming that the ratio of assumption-violating pixels is not big enough to greatly deteriorate the supervision signal.
	
	We compare two content-aware loss functions. The first one is proposed in \cite{zhou2017unsupervised}. The predicted mask is called the explainability map. Its elements $E_k$ are bounded between zero and one by a Sigmoid activation. $E_k$ indicates the network’s belief in how much the assumptions are satisfied for the $k$th pixel of $\mathcal{I}_{p}$. The pixel-wise loss is weighted by $E_k$ as shown in \eqref{Exp_loss}.
	A regularization term $\mathcal{L}_{\rm reg.}(E_k)$ encourages non-zero $E_k$ by minimizing the cross-entropy loss with $1.0$ so as to prevent the network to minimize the loss by outputing small values for all $E_k$.
	If the network predicts the $k$th pixel to meet the assumptions well, the value of $E_k$ would be close to $1.0$ and $\mathcal{L}(\bm{\mathcal{I}}_{4,k})$ would be fully minimized. On the contrary, $\mathcal{L}(\bm{\mathcal{I}}_{4,k})$ would be ignored if $E_k$ is close to zero. 
	\begin{equation}{\label{Exp_loss}} 
		\mathcal{L}_{\rm CA, Exp.} = \frac{1}{|V|}\sum_{k\in V} E_k \cdot \mathcal{L}(\bm{\mathcal{I}}_{4,k}) + \lambda_{\rm reg.} \cdot \mathcal{L}_{\rm reg.}(E_k)
	\end{equation}
	
	Another approach considers the content-aware mask and homography transformation as parameters of a Laplacian probability distribution. The nature of the mask is an uncertainty map. This approach was adopted for structure from motion \cite{klodt2018supervising, yang2020d3vo} and optical flow estimation \cite{ilg2018uncertainty}. Given the Laplacian probability density function (PDF)
	\begin{equation}{\label{Lap_pdf}}
		p(x| \mu,b) = \frac{1}{2b}\,e^{\frac{-|x-\mu|}{b}},
	\end{equation}
	since we use $L$1 loss in \eqref{pixel_loss}, the term $|x-\mu|$ can be replaced by photometric matching loss $\mathcal{L}(\bm{\mathcal{I}}_{4,k})$ calculated from the homography prediction. 
	{The parameter $b$ in \eqref{Lap_pdf} is related to the variance $\sigma^2 = 2 b^2$ of the Laplacian distribution.
	The predicted mask is made of the $b_{k}$ that corresponds to the $k$th pixel of the photometric matching map $\bm{\mathcal{I}}_4$.}
	The learning objective is to maximize the PDF, \textit{i.e.}, minimize the NLL loss
	\begin{equation}{\label{NLL_Lap}}
		\mathcal{L}_{\rm CA, Lap.} = \frac{1}{|V|} \sum_{k\in V} \frac{\mathcal{L}(\bm{\mathcal{I}}_{4,k})}{b_{k}} + {\rm log} \, b_{k}.
	\end{equation}

	$b_{k}$ can be understood intuitively as the uncertainty of the indirectly predicted $\mathcal{L}(\bm{\mathcal{I}}_{4,k})$, \textit{i.e.} photometric matching uncertainty.
	From the perspective of the uncertainty of network prediction, $b_k$ encodes the
	predictive uncertainty induced by the content-related observation noise.
	If pixel $k$ potentially violates the assumptions and is too difficult for photometric matching, \eqref{NLL_Lap} allows the learning process to increase the value of $b_k$ to down-weight $\mathcal{L}(\bm{\mathcal{I}}_{4,k})$ and reduce the overall loss.
	${\rm log} \, b_{k}$ prevents $b_k$ to overgrow.
	
	\begin{table}[!htbp]
		\centering
		\setlength{\tabcolsep}{0.52mm}{
			\caption{Comparison of Content-Aware Homography Networks (CAHN)}
			\label{masks}
			\centering
			\begin{threeparttable}
				\begin{tabular} {c | c c c c c}
					\toprule
					Setups& No Mask& Lap.& Exp. ($\lambda_{\rm reg.}=$1e-3)& Exp. (2e-3)& Exp. (3e-3) \\
					\midrule 
					Avg. Exp.& -&  -&  0.383&  0.640& 0.779\\ 
					\midrule
					Avg. Error $\downarrow$& 0.8768&  0.8477&  0.8483&  0.8471& 0.8487\\ 
					\bottomrule 
				\end{tabular}
			\end{threeparttable}
		}
	\end{table}
	
	The content-aware networks are trained and tested on the MYNT Dataset. Except for the randomly initialized mask prediction decoder, all parameters are initialized by the Basic Model.
	{The average error of the predicted $\bm{f}_{\text{total, CA}}$ of each setup is shown in the 3rd row of Table \ref{masks}.}
	The 2nd column is the baseline network without mask prediction and trained by the original loss function \eqref{self_loss}. The network of the 3rd column is trained by \eqref{NLL_Lap}. The rest three columns correspond to networks trained by \eqref{Exp_loss} with different hyperparameters $\lambda_{\rm reg.}$. Their pixel-wise average values of the predicted explainability maps on the training set are listed in the 2nd row.
	A bigger value means that more photometric error is taken into account in the loss function. 
	
\begin{figure}[!hbpt] 
		\centering
		\makebox{
			\includegraphics[scale=0.251]{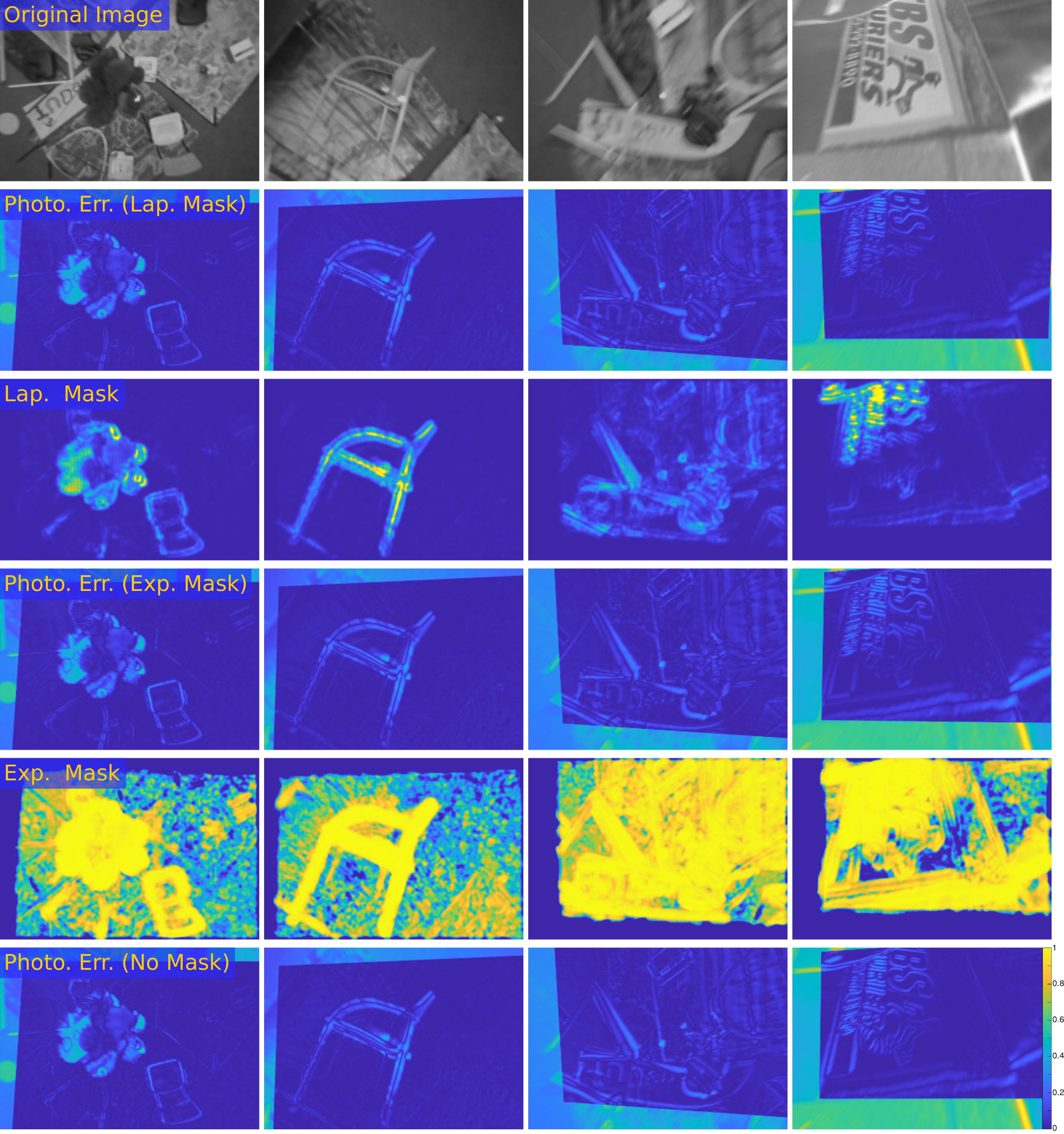}}
		\caption{From the top row to the bottom row: original image, photometric error maps of the network trained by \eqref{NLL_Lap}, uncertainty maps ($b_k$) in \eqref{NLL_Lap}, photometric error maps of the network trained by \eqref{Exp_loss}, explainability maps ($E_k$) in \eqref{Exp_loss} ($\lambda_{\rm reg.}=$2e-3), and photometric error maps of the baseline network. 
		{The photometric error map is made of $|\mathcal{I}_{p,k} - \widetilde{\mathcal{I}}_{c,k}|$, where $\widetilde{\mathcal{I}}_{c}$ is the warped ${\mathcal{I}}_{c}$ according to the predicted $\bm{f}_{\text{total}}$ and $k$ is the pixel index. Dark blue means a low error. The 5th row shows the maps of $1-E_k$. Pixels having small weight in the loss are in yellow, consistent with other rows.}
		The three columns on the left show example images from the MYNT Dataset. The object in the center of the leftmost column is an artificial plastic tree with leaves swaying due to the downwash. The rightmost column shows an image from the Basic Dataset.
		}
		\label{fig_mask}
	\end{figure}

	It is noticeable that the value of explainability map is sensitive to the manually set $\lambda_{\rm reg.}$, though the effect on prediction accuracy is small. Despite that 3-d structures distribute densely in the MYNT Dataset, the prediction accuracy of all setups is only slightly improved from the baseline. As shown in the 6th row of Fig. \ref{fig_mask}, the example photometric error maps of the baseline look very similar to the 2nd and 4th rows that are outputs of the content-aware networks. 
	The reason behind the minor contribution of content-aware learning can be that the original loss function \eqref{self_loss} minimizes the total photometric error, which drives the network to abandon the minority assumption-violating pixels. 
	
	As shown in Fig. \ref{fig_mask}, the uncertainty map (3rd row) is clear and corresponds well to the photometric error (2nd row) caused by non-homography image content. The explainability mask (5th row) is very noisy and prone to discount the textures because they cause bigger photometric errors than uniform regions. Besides, the sensitivity of the explainability mask toward $\lambda_{\rm reg.}$ may induce extra work of parameter tuning. Therefore, we believe that the uncertainty map trained by \eqref{NLL_Lap} is the better choice for content-aware learning. 

	{The 2nd and 3rd rows of Fig. \ref{fig_mask} show the positive correlation between the predicted uncertainty maps and the photometric error maps. Photometric error can be caused by non-homography image content and inaccurate homography transformation. 
	When obvious non-homography pixels exist, we can observe that most pixels with high predicted uncertainty fall on 3-d structures as shown in the three columns on the left. 
	When the scene is perfectly planar, as shown in the rightmost column, the non-zero uncertainty predictions are totally caused by the homography prediction error.
	So content-aware mask is not ideal for semantic plane segmentation. Its only duty is to down-weight the non-homography pixels in training.}
	
	\section{Uncertainty Estimation} \label{uncertainty}
	
	In Subsection \ref{literature_uncertainty}, we introduced practical approaches for estimating predictive uncertainty and empirical uncertainty. {In this section, they are implemented for uncertainty estimation of the homography network. We gain the knowledge of their uncertainty estimation quality, effects on prediction accuracy, and additional time consumption.}
	Uncertainty-Aware Homography Networks (UAHN) in this section are trained and evaluated on the Basic Dataset. Content-aware learning is not involved.
	
	\subsection{Configurations} 
	
	The correlations between the uncertainty of network outputs
	are often neglected in practice.
	The covariances between the pixel-wise predictions are not considered in monocular depth and semantic segmentation \cite{kendall2017uncertainties, poggi2020uncertainty, klodt2018supervising}. 
	The uncertainty of $u$ and $v$ components of an optical flow vector are separately estimated in \cite{ilg2018uncertainty}.
	As for pose prediction \cite{wang2018end, kaufmann2019beauty, li2020towards, wagstaff2022self}, the six elements are modelled as independent of each other.
	In this work, we neglect the covariances as well and leave them for potential future works.
	A scalar variance is estimated for each element of the 8-d mean prediction of homography transformation $\bm{f}_{\text{total}}$. 

	As introduced before, our network has four cascaded blocks that infer from their own inputs. It is not necessary for all of them to estimated the prediction uncertainty. The uncertainty of $\bm{f}_{4}$ estimated by the last (4th) block is enough to obtain the uncertainty of $\bm{f}_{\text{total}}$ by the following way.
	The 4th block infers from $\widetilde{\mathcal{I}}_{c,3}$ and ${\mathcal{I}}_{p}$ and outputs the mean prediction $\bm{f}_{4, j}$ and the variances $\sigma_{u,j}^2$, $\sigma_{v,j}^2$ of the endpoint of $\bm{f}_{4, j}$ in the 2-d image plane of $\widetilde{\mathcal{I}}_{c,3}$. $j$ indexes over the four corner pixels. $\bm{\Sigma}_{4,j}$ is a 3-by-3 diagonal matrix whose diagonal elements are $\sigma_{u,j}^2$, $\sigma_{v,j}^2$ and zero. 
	The coordinate of the pixel in ${\mathcal{I}}_{c}$ that has the same intensity as the endpoint of $\bm{f}_{4, j}$ can be obtained by \eqref{linear_transfor_f_vec}. Thus the variance of this pixel coordinate, \textit{i.e.}, the variance $\bm{\Sigma}_{\text{total}, j}$ of $\bm{f}_{\text{total}, j}$, is calculated as
	\begin{equation}{\label{var_trans}}
		\lambda^2 \bm{\Sigma}_{\text{total}, j} = \bm{H}_{\text{integ.},3} \cdot \bm{\Sigma}_{4,j} \cdot \bm{H}_{\text{integ.},3}^{T}.
	\end{equation}
	Note that the $\lambda$ here has the same value as the $\lambda$ in \eqref{linear_transfor_f_vec}. 
	Based on the above explanation, the first three blocks of UAHN stay the same as the Basic Model. Only the 4th block is modified for uncertainty estimation.
	
	$\bm{\Sigma}_{4,j}$ is a diagonal matrix but $\bm{\Sigma}_{\text{total}}$ has non-zero non-diagonal elements because of the matrix multiplication. These non-diagonal elements are two orders of magnitude smaller than the diagonal elements in general. So we neglect them and form up the error-variance pair for evaluation by the 2-d error of $\bm{f}_{\text{total}, j}$ and the first two diagonal elements of $\bm{\Sigma}_{\text{total},j}$.
	In this way, a testing image pair has eight error-variance pairs.
	
	One of the metrics to evaluate the quality of uncertainty estimation is Area Under the Sparsification Error (AUSE). It derives from the ``sparsification plot" that reflects how well high errors and high uncertainty coincide.
	To form a sparsification plot, error-variance pairs are descendingly sorted according to variance. Pairs with the highest variances are removed gradually. If the variances and errors coincide well, the average error of the remaining pairs should decrease while we are removing the data. In contrast, there would be little change in the average error if the variance does not correlate with the error. The ideal sparsification \textit{i.e., oracle sparsification}, is obtained by removing data pairs with the highest errors gradually. An example is the rightmost subplot of Fig. \ref{3_sigma}. The horizontal and vertical coordinates are respectively the ratio of removed data and the average error of remaining data.
	
	The difference between the sparsification formed up by the estimated variances and the ideal sparsification reflects the quality of variance estimation. A sparsification error curve is calculated by subtracting the ideal sparsification from the estimated one. 
	AUSE is the area of the region below the error curve. A lower AUSE means better variance estimation.
	In practice, to reduce the computation, data pairs are removed in batches. 
	In this article, we remove ten pairs at each step to get a data point of the sparsification curve. 
	An AUSE value shown later is the sum of the vertical axis coordinates of all the data points of the sparsification error curve.
	
	AUSE only reflects the relative values among the variances without showing how well their values reflect the values of the errors. For example, for three errors, 1, 2, and 3, the corresponding variances estimated by two approaches are 0.1, 0.2 0.3, and 10, 20, 30, respectively. In this case, their sparsification plots are the same but obviously the former approach underestimates and the latter overestimates the uncertainty.
	So we use another metric as a complement. It is the percentage of the testing errors falling into the three standard deviations ($3 \sigma$) interval, abbreviated as ``Inside Rate". A low Inside Rate means that the uncertainty is underestimated.
	
	\subsection{Model Distillation for Predictive Uncertainty} \label{pred_uncertainty}
	
	As introduced in Subsection \ref{mask}, an additional decoder network predicts the photometric matching uncertainty per image pixel, with the purpose of content-aware learning.
	Here we shift the focus to the predictive uncertainty of the homography transformation parameterized as the 8-d corner flow ${\bm f}_4$.
	The approach proposed in \cite{nix1994estimating} is adopted. 
	A subnetwork of two fully-connected (FC) layers is added to the 4th block to infer the predictive uncertainty from input. It has the same architecture and input tensor as the layers predicting the mean values.
	The outputs are eight logarithmic variances, ${\rm log} \, \sigma_{u,j}^2$ and ${\rm log} \, \sigma_{v,j}^2$. 
	The training loss is the NLL loss
	\begin{equation}{\label{NLL_Gaus}}
		\mathcal{L}_{\rm Gaus.} = \sum_{n=1}^{8} \frac{1}{2\sigma_n^2(\bm{\mathcal{I}}_3)}||t_{n}-\mu_n(\bm{\mathcal{I}}_3)||^2 + \frac{1}{2} {\rm log} \, \sigma_n^2(\bm{\mathcal{I}}_3).
	\end{equation}
	$t_{n}$ denotes the learning target of the mean value. $\mu_n$ and $\sigma_n^2$ are respectively the means and variances inferred from the input $\bm{\mathcal{I}}_3$.
	$n$ indexes over the elements of ${\bm f}_4$.
	
	Since we aim to build a self-supervised pipeline, ground-truth $t_{n}$ is not available. 
	Inspired by the Self-Teach scheme proposed in \cite{poggi2020uncertainty}, the pseudo label $t_{n}$ can be generated by a network trained in self-supervised fashion. A student network predicting both mean and variance can be trained by \eqref{NLL_Gaus} under the supervision of the trained teacher network that outputs only the mean predictions.
	The student is trained to imitate the teacher by outputting $\mu_n$ closer and closer to $t_{n}$. 
	The predictive variance $\sigma_n^2$ learns to capture how good is the imitation.
	Thus $\sigma_n^2$ only reflects the imitation error $||t_{n}-\mu_n||^2$ instead of the true error of $\mu_n$ \textit{w.r.t.} the ground truth. 
	
	The Basic Model has decent accuracy and thus is an option for the teacher network.
	We referred to it as Self-Teach, same as \cite{poggi2020uncertainty}.
	Besides, we propose an enlarged version of the Basic Model called the Master Model. It has six network blocks in total. The first three are the same as the Basic Model. The following three blocks have the same architecture as the 4th block of the Basic Model. They together can be treated as a more capable ``last block". In the refining training of the Master Model, we initialized the last three blocks by the parameters of the 4th block of the Basic Model. A small improvement in accuracy was achieved and the final testing average error is 0.144 pixels, better than the Basic Model (0.275). The scheme using the Master Model as teacher is called Master-Teach. 
	
	Self-Teach and Master-Teach are compared in Table \ref{variance_comparison_supervision}. To gain more insight, we also trained a student network supervised by ground-truth $t_{n}$, abbreviated as GT-Teach. $t_{n, {\text{GT}}}$, \textit{i.e.} $\bm{f}_{4,{\text{GT}}}$, is calculated from $\bm{f}_{\text{total, GT}}$ and $\bm{H}_{\text{integ.},3}$ that is predicted by the first three blocks.
	The 4th blocks of all the student networks in Table \ref{variance_comparison_supervision} are randomly initialized before training. 
	When a training epoch has been finished, the average imitation error is calculated on the validation set. 
	The set of network parameters achieving the smallest average imitation error are recorded for testing.
	
	\begin{table}[!htbp]
		\centering
		\setlength{\tabcolsep}{0.52mm}{
			\caption{Comparison of Supervision Signals for Predictive Uncertainty}
			\label{variance_comparison_supervision}
			\centering
			\begin{threeparttable}
				\begin{tabular} {m{44pt}<{\centering}m{35pt}<{\centering}m{48pt}<{\centering}m{36pt}<{\centering}m{28pt}<{\centering}m{39pt}<{\centering}}
					\toprule
					Supervision& Avg. Error (pixel) $\downarrow$& Avg. Imitation Error (pixel) &Avg. Var.\tnote{$\star$} (pixel) & AUSE $\downarrow$ & Inside Rate (\%) ($3 \sigma$) $\uparrow$ \\
					\midrule
					
					GT-Teach&  0.454&  0.489&  16.66& 370.0& \textbf{97.63}\\ 
					Master-Teach& 0.428&  0.336&  7.69& \textbf{357.3}& 86.96\\ 
					Self-Teach&  \textbf{0.408}& {0.208}& {1.83}& 565.9& 79.08\\ 
					
					\bottomrule
				\end{tabular}
				\begin{tablenotes}
					\footnotesize
					\item[$\star$] The networks predict rare unreasonably big variances. The averages are calculated after removing the 0.1\% biggest values.
				\end{tablenotes}
			\end{threeparttable}
		}
	\end{table}
	
	\begin{figure*}[!hbpt]
		\centering
		\makebox{
			\includegraphics[scale=0.40]{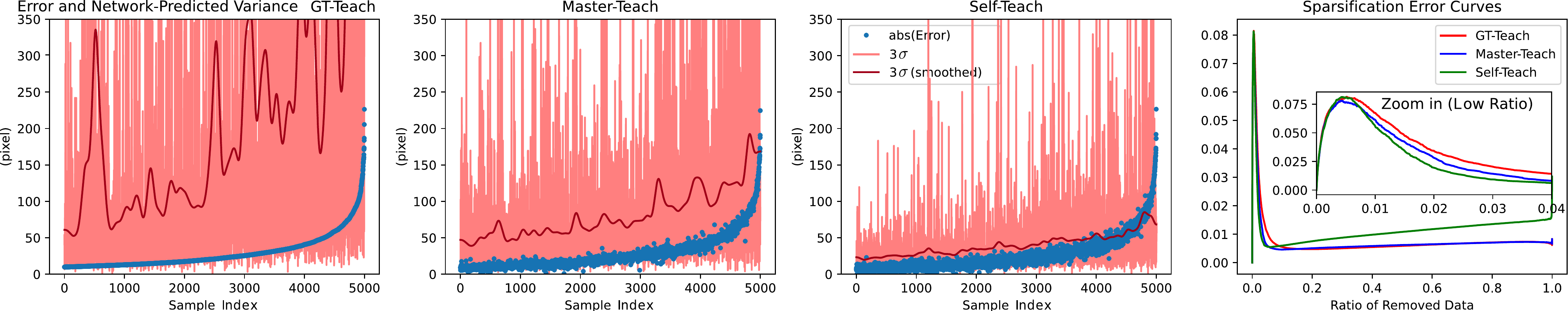}}
		\caption{Comparison of different approaches to learning predictive uncertainty. Left three subplots: three times of the predictive standard deviations $\sigma$ and the corresponding prediction errors (absolute values) of GT-Teach, Master-Teach, and Self-Teach, sorted according to the errors of GT-Teach. The 5,000 samples with the \textbf{biggest testing errors} are shown. Because $3 \sigma$ is very noisy, a Gaussian filter is adopted to produce the smoothed curves (in dark red) that allow more intuitive views. Rightmost subplot: comparison of the sparsification error curves of the three supervision schemes.}
		\label{variance_supervision_compare}
	\end{figure*}
	
	Table \ref{variance_comparison_supervision} shows that GT-Teach has the lowest prediction accuracy but the highest Inside Rate and average variance. The AUSE of Master-Teach is the lowest but its advantage over GT-Teach is small. For all other metrics, Master-Teach achieved the middle places.
	The smallest imitation error and variance indicate that the student model imitates the teacher best in the Self-Teach scheme. But the AUSE and Inside Rate
	tell us the predictive uncertainty of Self-Teach is the poorest. The low Inside Rate and average variance show that the uncertainty is underestimated. 

	To visualize the comparison better, we plot the prediction errors and predictive variances in Fig. \ref{variance_supervision_compare}. 
	{For most testing samples, their error and predictive variance are both small. Here we show the 5,000 error-variance pairs with the biggest errors. 
	Inaccurate predictions like them are dangerous for VIO if the corresponding high variances are not correctly predicted.
	The error-variance pairs from different supervision schemes are aligned by data indexes and sorted according to the errors of GT-Teach.
	In this way, the data points with the same index in the three subplots correspond to the same element of the corner flow of the same image pair. 
	We can observe that the three schemes have similar errors for the same testing sample, consistent with the similar average errors in Table \ref{variance_comparison_supervision}. }
	$3 \sigma$ grows with error as a general trend. 
	$3 \sigma$ of GT-Teach is the noisiest and biggest. In contrast, Self-Teach has the smallest $\sigma$ that is sluggish toward the increasing error and tends to underestimate especially the big errors.
	The sparsification error curves shown in the rightmost subplot have peaks at a very low ratio, which means that, statistically, the quality of predictive variance is poor when the prediction error is big. For most of the testing data, predictive variance is effective, as evidenced by the low sparsification error curves. 
	
	In training, the prediction error of a student network is caused by two factors, the imitation error $||t_{n}-\mu_n||^2$ and the prediction error of the teacher. As mentioned before, $\sigma_n^2$ can only capture the imitation error. 
	So when the imitation errors are big and the teacher errors are small, $\sigma_n^2$ well reflects the prediction errors. Conversely, when imitation error is small but the teacher predictions are inaccurate, $\sigma_n^2$ keeps a small value and becomes almost irrelevant to the student prediction error. 
	Master-Teach and Self-Teach respectively correspond to the former and latter cases above. Thus Master-Teach produces better predictive uncertainty.

	For a sparsification curve, when the ratio of removed data goes higher, remaining data pairs have smaller $\sigma_n^2$.
	As shown in the rightmost subplot of Fig. \ref{variance_supervision_compare}, Self-Teach has an increasing sparsification error curve, which indicates that a smaller $\sigma_n^2$ coincides worse with the actual error. 
	The reason for a small $\sigma_n^2$ can be that the student network is confident that its mean value prediction is close to the teacher network that supervised it in training. 
	In this case, the student prediction error is close to the unknown teacher error that is not reflected by $\sigma_n^2$.
	
	\begin{figure}[b]
		\centering
		\makebox{
			\includegraphics[scale=0.375]{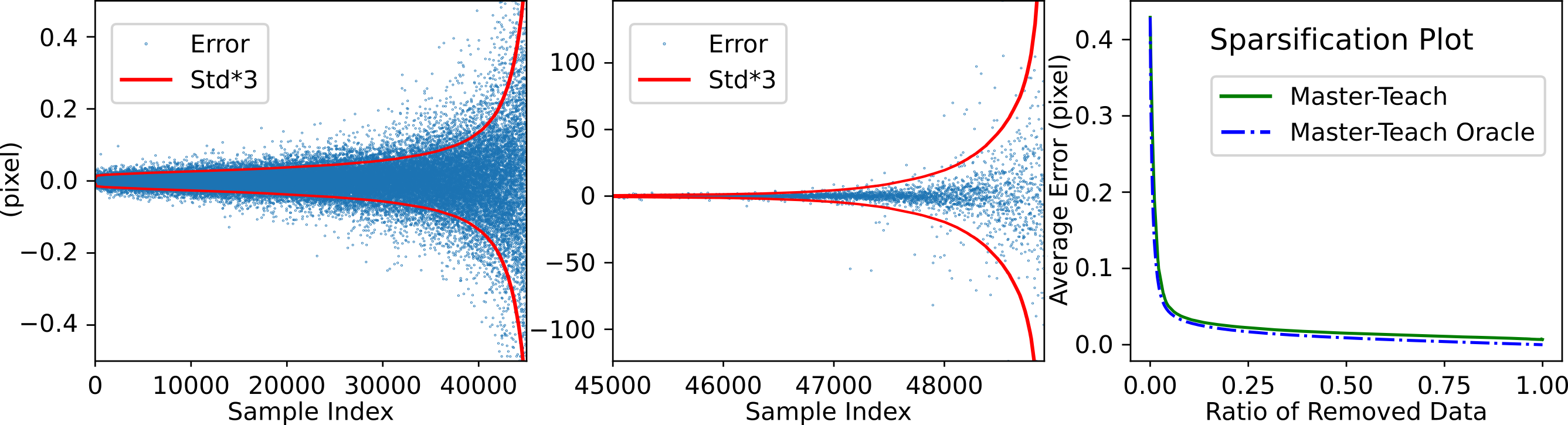}}
		\caption{Predictive uncertainty of the randomly initialized Master-Teach model (1st row of Table \ref{variance_comparison_init}). Error-variance pairs of the testing set are sorted according to the variances. To avoid overly dense data points in plots, we show one point every ten pairs. Small and big variances are shown respectively in the left and middle subplots with different ranges of $y$-axis for better visualization. 
		}
		\label{3_sigma}
	\end{figure}
	
	Fig. \ref{3_sigma} diagrams how the predictive variances cover the prediction errors in a different view from Fig. \ref{variance_supervision_compare}. The left subplot shows around 90\% of the testing data. Most data points fill up the area between $- 3 \sigma$ and $3 \sigma$. The local Inside Rate is 86.02\%. In the middle subplot, it is noticeable that the variances are much bigger for the remaining 10\% of the data. Although the distribution of the errors becomes broader, but not as much as $3 \sigma$ increases, \textit{i.e.} the extent of overestimation grows with $3 \sigma$. The local Inside Rate is 95.38\%.
	As observed in the sparsification plot (right subplot of Fig. \ref{3_sigma}), when the data pairs with the biggest predictive variances (less than 5\% of the total) are removed, the average error drastically drops to less than 0.1 pixels.
	It indicates that, for most testing samples, the Master-Teach student network achieves high prediction accuracy. The uncommon outliers can be revealed by the big predictive variances.
	The above results corroborate that the quality of Master-Teach predictive uncertainty is generally satisfactory. All the uncertainty-aware networks in the rest of this article are trained in Master-Teach scheme.
	
	\begin{table}[!htbp]
		\centering
		\setlength{\tabcolsep}{0.52mm}{
			\caption{Comparison of Different Initializations}
			\label{variance_comparison_init}
			\centering
			\begin{tabular} {m{32pt}<{\centering}m{38pt}<{\centering}m{32pt}<{\centering}m{41pt}<{\centering}m{27pt}<{\centering}m{25pt}<{\centering}m{32pt}<{\centering}}
				\toprule
				Conv. Layers& FC Layers (mean value)& Avg. Error (pixel) $\downarrow$ & Avg. Imitation Error (pixel) &Avg. Var. (pixel) & AUSE $\downarrow$& Inside Rate (\%) ($3 \sigma$) $\uparrow$ \\
				\midrule
				
				Random& Random&  0.428&  0.336&  7.69& \textbf{357.3}& \textbf{86.96}\\ 
				Basic& Random&  0.376&  0.282&  5.98& 400.0& 57.05\\ 
				Basic& Basic&  \textbf{0.352}&  {0.258}&  4.37& 387.9& 62.65\\ 
				
				\bottomrule
			\end{tabular}
		}
	\end{table}
	
	Besides random initialization, the student network can be initialized with the trained parameters of the Basic Model.
	It is clearly shown in Table \ref{variance_comparison_init} that initializing both convolutional layers and FC mean prediction layers (3rd row) with the Basic Model is better than convolutional layers alone (2nd row).
	Random initialization (1st row) has the best predictive uncertainty and worst prediction accuracy. 
	The high prediction accuracy of the 3rd row comes from the initial parameters. They also make the imitation error smaller.
	Thus the predictive variances are smaller and reflect the prediction errors less well, leading to worse predictive uncertainty.

	\subsection{Empirical Uncertainty} \label{empirical_uncertainty}

	\begin{figure*}[b]
		\centering
		\makebox{
			\includegraphics[scale=0.48]{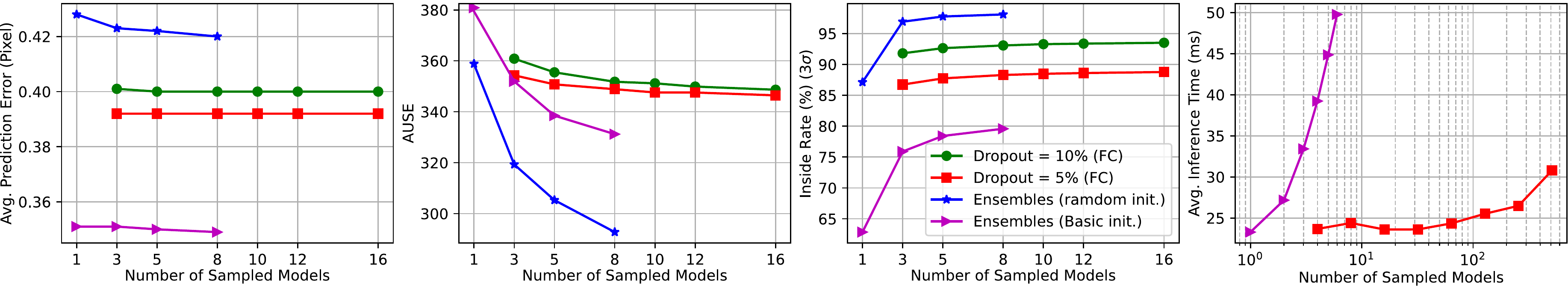}}
		\caption{Comparison of approaches for empirical uncertainty in terms of prediction accuracy, uncertainty estimation quality (AUSE and Inside Rate), and inference time consumption. The $x$-axis for deep ensembles is the number of independent network models. For MC-Dropout, the $x$-axis is the number of forward passes. Deep ensembles indicated by blue stars and magenta triangles correspond respectively to the 1st and 3rd row of Table \ref{variance_comparison_init}. The inference time was measured on a TX2 processor running network inference in a C++ environment. We show the inference time of one of the two networks using MC-Dropout because the other has the same in theory. Same for the deep ensembles.
		}
		\label{ensemble_comparison}
	\end{figure*}
	
	Deep ensembles \cite{lakshminarayanan2017simple} and MC-Dropout \cite{gal2016dropout} are implemented on the student networks with predictive uncertainty.
	They both require multiple forward passes to get the samples from the distributions of network parameters. The variance is calculated empirically from the outputs of the forward passes. 
	The total variance $\sigma^2_n$ of the $n$th element of ${\bm f}_4$ is shown in \eqref{comb_sigma} as the sum of the empirical variance of the mean value predictions $\mu_{m,n}$ and the average of predictive variances $\sigma^2_{m,n}$. $m$ indexes over the network model samples.
	\begin{equation}{\label{comb_mu}}
		\mu_n = \frac{1}{M}\sum_{m=1}^{M} \mu_{m,n}
	\end{equation}
	\begin{equation}{\label{comb_sigma}}
		\begin{aligned} 
			\sigma^2_n = &\sigma^2_{n,\text{pred.}} + \sigma^2_{n,\text{emp.}} \\
			\sigma^2_{\text{pred.}} = \frac{1}{M}\sum_{m=1}^{M} \sigma^2_{m,n},& \quad \sigma^2_{\text{emp.}} = \frac{1}{M}\sum_{m=1}^{M} (\mu_{m,n}-\mu_{n})^2
		\end{aligned}
	\end{equation}
	
	The idea of deep ensembles is to train $M$ network models independently as the samples. Training the models with different bootstrapped subsets of the training data enhances independence. But meanwhile, less training data harms the prediction accuracy. We follow the practice of \cite{lakshminarayanan2017simple} that using the entire training set for every model, assuming random initialization along with random shuffling of training data produce sufficient independence. 
	We trained eight independent models respectively for the 1st and 3rd initialization schemes in Table \ref{variance_comparison_init}. 
	An ensemble combines eight models at most because the increasing time consumption makes it impossible for real-time inferencing on a mobile processor.
	
	For MC-Dropout, we implement two schemes and two dropout rates. One scheme randomly initializes all parameters and performs dropout before all layers. The other initializes the convolutional layers with the parameters of the trained Basic Model. Following the practice of \cite{kendall2016modelling}, that is deploying dropout only before layers that are randomly initialized, dropout is only effective before FC layers.
	
	The above-introduced schemes are compared in Fig. \ref{ensemble_comparison} by four metrics.
	We find that performing dropout before all layers leads to much worse accuracy and AUSE, besides long inference time. So it is eliminated without being shown.
	The leftmost subplot shows that increasing the number of sampled network models only slightly improves prediction accuracy. 
	While, the AUSE values shown in the 2nd subplot from left vary with the number of samples significantly, especially for deep ensembles. The same trend is observed in Inside Rate (3rd subplot). An ensemble of three network models has significantly better uncertainty estimation than a single one. In contrast, more forward passes of MC-Dropout networks produce relatively smaller improvements.
	The higher dropout rate (10\%) performs worse than the lower one (5\%) in terms of both accuracy and AUSE, while better in Inside Rate.

	As for the two initialization schemes, random initialization has better and bigger predictive uncertainty as shown before. 
	Thanks to more randomness in network parameters, random initialization in theory has better and bigger empirical uncertainty as well and thus has better overall uncertainty estimation. As shown in Fig. \ref{ensemble_comparison}, the two initialization schemes are respectively advantageous in prediction accuracy and uncertainty estimation quality.
	
	For deep ensembles, the time consumption increases significantly with the number of sampled networks. We failed to find a way in our current implementation to speed up, though the samples of the 4th network block are independent and, in theory, can run in parallel.
	Due to the real-time requirement of VIO, we only consider the ensembles of less than three network samples. 
	For MC-Dropout, instead of inferencing multiple times temporally serially, the intermediate tensors can be duplicated along the batch dimension before dropout to obtain the same effect. The time consumption increases insignificantly as shown in the rightmost subplot of Fig. \ref{ensemble_comparison}.
	Based on the overall consideration of the four metrics, we select three candidates
	that act as the VIO vision frontend and are compared in terms of the resulting VIO accuracy in Subsection \ref{Variants_CUAHN}. They are
	1) the ensemble of randomly initialized models (indicated by the blue stars in Fig. \ref{ensemble_comparison}), 2) the ensemble of models initialized by the Basic Model (magenta triangles), and 3) 5\% dropout before FC layers with 16 forward passes (red squares). 
	
	{\textit{Supplementary Document} shows the magnitudes of $\sigma^2_{\text{emp.}}$ and $\sigma^2_{\text{pred.}}$ and their correlation for interested readers. }
	
	\section{Visual-Inertial Odometry}
	
	\subsection{Homography-Network-based Vision Frontend}
	
	We have introduced how to train CAHN (Section \ref{homo_network}) and UAHN (Section \ref{uncertainty}). In the following, we describe the way of combining both to get a content-and-uncertainty-aware homography network (CUAHN), and how it acts as the vision frontend of a VIO system.
	
	{As discussed in the previous section, Master-Teach is a good choice for the student network to learn the predictive uncertainty. 
	To gain higher accuracy through the robustness toward non-homography image content, we train the Master Model by the content-aware loss \eqref{NLL_Lap}, different from the Master Model in Subsection \ref{pred_uncertainty} that minimizes the photometric error of all the pixels.
	Three upsampling decoders are respectively attached to the last three blocks to predict the content-related photometric matching uncertainty maps, as introduced in Subsection \ref{mask}. The decoders share parameters. 
	In this way, each of the last three blocks has its predicted uncertainty map and the photometric matching map obtained from the integrated homography transformation prediction $\bm{H}_{\text{integ.},i}$. 
	The training loss is the sum of the content-aware losses of the three blocks.}
	
	It is important for the training set to have enough non-homography contents and also be generic. In this article, the six sequences with public available ground truth of UZH-FPV are used for evaluation. 
	We take the 6,070 image pairs from the rest four sequences without ground truth for training and name them the UZH-FPV training set. Together with the generic Basic Dataset, the aggregated dataset is used for training CUAHN. 
	
	As introduced in Section \ref{uncertainty}, an uncertainty-aware network estimates the 8×8 covariance matrix $\bm{R}_{\text{net.}}$ of the corner flow prediction. Theoretically, it should be used directly as the measurement noise covariance matrix $\bm{R}_{\text{meas.}}$ in the measurement update of EKF. But because the performance of the EKF is under the effects of noise matrices, it is better to have the freedom of tuning the measurement noise. 
	Thus we introduce a manually tuned scalar hyperparameter $k_{\text{var.}}$ to linearly scale $\bm{R}_{\text{net.}}$ as $\bm{R}_{\text{meas.}} = k_{\text{var.}} \cdot \bm{R}_{\text{net.}}$. In practice, it is easy to tune since the system is not very sensitive to $k_{\text{var.}}$.
	
	\subsection{EKF-based Backend} \label{ekf_backend}
	The VIO backend is a simple and very efficient EKF.
	IMU measurements drive the state propagation and network outputs drive the visual measurement updates. 
	To simplify the filter, we assume that a single plane is observed by the camera throughout the whole video and the plane is orthogonal to the gravity vector. These assumptions apply to many flight arenas, especially indoor environments.
	The origin of the world frame lies on the plane and the $z$-axis is parallel to the gravity vector as shown in Fig. \ref{mav_cam}.
	The EKF state vector is defined as:
	\begin{equation}{\label{state_def}}
		\bm{x} := \begin{bmatrix}
			\bm{p}, \bm{q}, \bm{v}, \bm{b}_a, \bm{b}_g, 
			\bm{f}_{j}
		\end{bmatrix},
		\quad j\in\{ul, bl, br, ur\}.
	\end{equation}
	$\bm{p}$ is the position of IMU relative to the origin of the world frame, expressed in the IMU frame.
	$\bm{q}$ is the Hamilton quaternion reflecting the relative rotation between the world frame and the IMU frame. $\bm{v}$ is the translational velocity of IMU expressed in the IMU frame.
	$\bm{b}_a$ and $\bm{b}_g$ are respectively the additive bias on accelerometer and gyroscope. 
	$\bm{f}_j$ indicates the optical flow vector of the $j$th corner pixel between two consecutive frames.
	The foot markers of $\bm{f}_j$ are respectively the abbreviations of upper left, bottom left, bottom right, and upper right. 
	
	As shown in \eqref{imu_reading}, IMU measurements are modelled as the sum of the desired actual value ($\hat{\bm{a}}$ and $\hat{\bm{\omega}}$), additive bias ($\bm{b}_a$ and $\bm{b}_g$), and white Gaussian noise ($\bm{w}_a$ and $\bm{w}_g$). 
	\begin{equation}{\label{imu_reading}}
		\bm{a}_m = \hat{\bm{a}} + \bm{b}_a + \bm{w}_a, \quad \bm{\omega}_m = \hat{\bm{\omega}} + \bm{b}_g + \bm{w}_g
	\end{equation}
	\begin{equation}{\label{state_prop}}
		\begin{aligned}
			\dot{\bm{p}} &= -[\bm{\hat{\omega}}]_\times \bm{p} + \bm{v} + \bm{w}_{\bm p},\\
			\dot{\bm{v}} &= -[\bm{\hat{\omega}}]_\times \bm{v} + \hat{\bm{a}} + \bm{R}(\bm q)^{-1}{\bm g},\\
			\dot{\bm{q}} &= \dfrac{1}{2}\bm{q} \otimes \begin{bmatrix} 0 \\ \hat{\bm{\omega}} \end{bmatrix}, \\
			\dot{\bm{b}}_a &= \bm{w}_{\bm{b}_a},\quad 
			\dot{\bm{b}}_g = \bm{w}_{\bm{b}_g}, \\
			\dot{\bm{f}}_j &= -(\bm{I} - (\bm{c}_j+{\bm{f}}_j) \bm{e}_z^T)\bm{H}(\bm{c}_j+{\bm{f}}_j)
		\end{aligned}
	\end{equation}
	
	\eqref{state_prop} shows the IMU-driven state dynamics ($\dot{\bm{x}}$). 
	$[\bm{\hat{\omega}}]_\times$ represents the skew-symmetric matrix associated with $\bm{\hat{\omega}}$.
	$\bm{w}_{\bm p}$ is the process noise in position integration. 
	{$\bm{R}(\bm q)$ is a transformation function from $\bm q$ to \textit{SO3} rotation matrix that maps a vector expressed in the IMU frame to its expression in the world frame.}
	${\bm g}=[0, 0, g]^T$ is the gravity vector expressed in the world frame.
	$\otimes$ denotes quaternion product. We utilize the techniques introduced in \cite{sola2017quaternion} for quaternion-related calculation. 
	The propagation of $\bm{f}_j$ is based on the continuous homography transformation. The formula derivation can be found in \cite{zhong2020direct}. 
	$\bm{c}_j$ stands for the 2-d coordinate of the $j$th corner pixel. It is a constant parameter calculated from the camera intrinsics. $\bm{I}$ is a 3×3 identity matrix. $\bm{e}_z = [0, 0, 1]^T$. 
	In our implementation, $\bm{f}_j$ and $\bm{c}_j$ are homogeneous coordinates in the camera frame instead of pixel coordinates, which means that they are expressed on the $z=1$ plane of the camera frame. So camera intrinsics are not needed in state propagation.

	$ \bm{H} \in \mathbb{R}^{3\times 3} $ relates the camera motion to the optical flow $\dot{\bm{f}}_j$. It is known as the continuous homography matrix:
	\begin{equation}{\label{continue_H}}
		\bm{H} = \lbrack \hat{\bm{\omega}}_c \rbrack_\times+\dfrac{1}{d_c}\bm{v}_c\bm{\mu}_c^T
	\end{equation}
	where
	\begin{equation}{\label{iterms_continue_H_2}}
		\begin{aligned}
			\hat{\bm{\omega}}_c &= \bm{R}_{CI} \bm{\hat{\omega}}, \\
			\bm{v}_c &= \bm{R}_{CI}(\bm{v}+[\bm{\hat{\omega}}]_\times \bm{t}_{IC}), \\
		\end{aligned}
	\end{equation}
	and
	\begin{equation}{\label{iterms_continue_H_1}}
		\begin{aligned}
			\bm{\mu}_c^T &= \bm{R}_{CI}\bm{R}^{-1}(\bm q)\bm{e}_z, \\
			d_c &= -\bm{e}_z^T \bm{R}(\bm q) (\bm{p}+\bm{t}_{IC}). \\
		\end{aligned}
	\end{equation}
	$\hat{\bm{\omega}}_c$ and $\bm{v}_c$ are respectively the angular and translational velocity vectors of the camera expressed in the camera frame. $\bm{\mu}_c$ is the normal vector of the plane expressed in the camera frame. Based on our assumption, it has the same direction as the gravity vector. $d_c$ is the distance from the camera to the plane.
	We define the $z$-axis to be downward as shown in Fig. \ref{mav_cam}. In the cases where the $z$-axis points up, minus signs should be added to the right of the equal signs in \eqref{iterms_continue_H_1}.
	
	The visual measurement of $\bm{f}_j$ is modelled as
	\begin{equation}{\label{meas_eq}}
		{\bm z}_{j, t} = \bm{f}_{j, t|t-1} + \bm{w}_{j, t}
	\end{equation}
	where $\bm{f}_{j, t|t-1}$ is the \textit{a priori} estimation of $\bm{f}_{j}$ propagated by \eqref{state_prop}. ${\bm z}_{j,t}$ is the mean value prediction of the whole homography transformation from the network. When $\bm{f}_{t|t-1}$ is used for pre-warping as shown in Fig. \ref{system_overview}, the network predicts a part of the transformation and ${\bm z}_{j,t}$ is the combination of the network prediction and $\bm{f}_{t|t-1}$.
	$\bm{w}_{j, t}$ is the measurement noise. The covariance matrix of $\bm{w}_{j, t}$ is $\bm{R}_{\text{meas.}} = k_{\text{var.}} \cdot \bm{R}_{\text{net.}}$. 		
	Note that network outputs are in pixels. So ${\bm z}_{j,t}$ and $\bm{R}_{\text{net.}}$ are required to be scaled by the camera intrinsics (focal length) to convert to the homogeneous coordinates in the camera frame, the coordinate system same as $\bm{f}_{j, t|t-1}$.
	
	$\bm{f}_j$ is a temporary state reflecting the transformation between two consecutive frames. It has been propagating from zero since the acquisition of the last frame.
	When a new frame is available, the network inferences from the newest two frames and the difference between the propagated prior ${\bm f}_{j,t|t-1}$ and ${\bm z}_{j,t}$ acts as the measurement residual in EKF update. After updating, ${\bm f}_{j}$ and its corresponding elements in the covariance matrix of the state vector are reset to zeros.
	
	\section{Evaluation} \label{evaluation}
	We first compare the proposed VIO with open-sourced SOTA VIO approaches, followed by an ablation study. Then, a generic and efficient variant UAHN-VIO is demonstrated competent for feedback control of autonomous MAV flight in an unseen test environment. Lastly, we compare CUAHN-VIO with a feature-point-based VIO approach MSCKF \cite{geneva2020openvins, li2013high}, focusing on analyzing the processing latency and robustness toward fast motion. 
	
	The evaluation is mainly by means of the six indoor 45-degree downward-facing sequences trajectories from a public MAV dataset UZH-FPV \cite{delmerico2019we}. It is known for its high flight speed and big optical flow.
	Another dataset is recorded in autonomous MAV flights by the same hardware as the MYNT Dataset. It features frequent significant motion blur and is utilized in robustness evaluation. 
	
	As is common in VIO studies, the root-mean-square error (RMSE) of absolute translation errors (ATE) acts as the metric for VIO accuracy. We utilize an open-sourced tool \cite{zhang2018tutorial} for the calculation. The estimated trajectory and the ground truth are aligned by the 4-DoF yaw-only rigid body transformation (one DoF rotation plus 3-d translation, \textit{posyaw}) corresponding to the four unobservable DoFs for VIO \cite{zhang2018tutorial}.
	It reveals how well the scale of the estimated trajectory matches the metric scale. The \textit{Sim3} alignment widely used by other works cannot.
	
	\subsection{Comparison of Accuracy with SOTA VIO Approaches} \label{Comparison_SOTA}
	
	\begin{table}[!htbp]
	\centering
	\setlength{\tabcolsep}{0.52mm}{
		\caption{Comparison with SOTA VIO Approaches. \textbf{Bold}
			Represents the Best and \underline{Underline} Represents the Best of SOTA Approaches.}
		\label{SOTA_VIO}
		\centering
		\begin{threeparttable}
			\begin{tabular} {m{55pt}<{\centering}m{28pt}<{\centering}m{28pt}<{\centering}m{28pt}<{\centering}m{28pt}<{\centering}m{28pt}<{\centering}m{28pt}<{\centering}}
				\toprule
				\multirow{2}{*}{VIO}&
				\multicolumn{6}{c}{RMSE (meter) of Absolute Translation Errors (ATE) $\downarrow$}\\
				& Seq. 2 & Seq. 4 & Seq. 9 & Seq. 12 & Seq. 13 & Seq. 14 \\
				\midrule
				
				UAHN (6-3)& \textbf{0.3371}& 0.3139& 0.3392& 0.5837& {0.4066}& 1.7905 \\
				CUAHN (6-4)& 0.3479& {0.3138}& 0.3214& 0.5843& 0.4095& 1.7790 \\
				CUAHN (4-4)& 0.3475& \textbf{0.2739}& \textbf{0.3200}& \textbf{0.5826}& \textbf{0.3985}& 1.7903 \\ 
				\midrule
				
				OpenVINS \cite{geneva2020openvins}& \underline{0.3438}& 0.3937& \underline{0.3772}& 0.6252& 0.5542& \underline{\textbf{1.7392}}\\
				
				LARVIO \cite{qiu2020lightweight}& 1.0584& 0.8085& 0.5069& 0.8100& 0.8370& 2.0767 \\
				
				MSCKF \cite{li2013high, geneva2020openvins}& 0.3718& \underline{0.3704}& 0.4189\tnote{\dag}& 0.6347& 0.5424& 1.7405\\
				
				ROVIO \cite{bloesch2017iterated}& 0.9175\tnote{\dag}& 0.4233& 0.7837\tnote{\dag}& 0.6234& \underline{0.4217}& 1.8286\tnote{\dag}\\
				
				VINS-Fusion \cite{qin2017vins}& 0.4040& 0.4533\tnote{\dag}& 0.6439& \underline{0.6021}& 0.4544& 1.7988\tnote{\dag}\\
				
				\bottomrule
			\end{tabular}
			\begin{tablenotes}
				\item[\dag] Without online calibration.
			\end{tablenotes}
		\end{threeparttable}
	}
	\end{table}
	
	Based on the ablation study shown later, three setups of CUAHN-VIO (6-3, 6-4, and 4-4 in Table \ref{CUAHN_variants_compare}) are selected to compare with open-sourced SOTA VIO approaches in Table \ref{SOTA_VIO}. The numbers in the 1st row are the sequence numbers. The maximum speeds of the sequences in meters per second (m/s) are shown in the brackets following the sequence numbers in the 1st row of Table \ref{CUAHN_variants_compare}.
	We used a laptop computer to run the VIO approaches to guarantee no frame was discarded because of slow processing.
	We tried to get as good as possible results from the SOTA approaches by tuning the parameters\footnotemark[\value{footnote}], \textit{e.g.} IMU noise density and the starting time of the data sequences. 
	For approaches having the function of online calibration,
	we tried both with and without this function and put the better results in the table. ORB-SLAM3 \cite{campos2021orb} was also tried but it failed to initialize the map or keep tracking it on any sequence. 
	We do not compare with learning-based VIO because we failed to find an open-sourced one that can run on the evaluation dataset without (re)training.
	
	For five sequences out of six, the smallest errors are achieved by UAHN or CUAHN. For Seq. 4 and 9, the advantage is relatively obvious. In general, the proposed VIO rivals the SOTA approaches. 
	
	\begin{table*}[!htbp]
	\centering
	\setlength{\tabcolsep}{0.52mm}{
		\caption{Evaluation of VIO Variants by Indoor 45-degree Downward-facing Sequences of UZH-FPV. \textbf{Bold} Represents the Best Result.}
		\label{CUAHN_variants_compare}
		\centering
		\begin{threeparttable}
			\begin{tabular} {m{25pt}<{\centering}m{44pt}<{\centering}m{32pt}<{\centering}m{28pt}<{\centering}m{32pt}<{\centering}m{50pt}<{\centering}m{48pt}<{\centering}m{32pt}<{\centering}m{32pt}<{\centering}m{32pt}<{\centering}m{32pt}<{\centering}m{32pt}<{\centering}m{32pt}<{\centering}}
				
				\toprule
				\multirow{2}{*}{Number}&
				\multirow{2}{*}{Network}&
				\multirow{2}{*}{\makecell[c]{Initial-\\ization}}&
				\multirow{2}{*}{\makecell[c]{Num. of\\Blocks}}&
				\multirow{2}{*}{\makecell[c]{$\bm{R}_{\text{meas.}}$/\\$k_{\text{var.}}$}}&\multirow{2}{*}{\makecell[c]{Empirical\\Uncertainty}}&
				\multirow{2}{*}{\makecell[c]{Avg. Time \\ Consumption\tnote{$\star$} $\downarrow$}}& 
				\multicolumn{6}{c}{RMSE (meter) of Absolute Translation Errors (ATE) $\downarrow$}\\
				& & & & & & & 2 (6.97)& 4 (6.55)& 9 (11.23)& 12 (4.33)& 13 (7.92)& 14 (9.54)\\
				
				\midrule
				\rowcolor{lightyellow} 1-1& Basic& rand.& 4& 125.0& None& 20.755& 5.3298& 4.9616& 2.6323& 3.0957& 2.5465& 4.3532 \\ 
				\rowcolor{lightyellow} 1-2& CAHN& Basic& 4& 35.0& None& -& 3.7962& 5.5064& 2.6658& 3.3070& 2.8477& 7.8038 \\
				\rowcolor{lightyellow} 1-3& Basic& rand.& 3& 10.0& None& 19.658& 1.1526& 1.0022& 0.5777& 0.9379& 1.2572& \textbf{1.6455} \\
				\rowcolor{lightyellow} 1-4& CAHN& Basic& 3& 1.5& None& -& 1.1091& 1.5402& 0.4282& 0.7613& 0.7873& 1.6754 \\
				
				\midrule
				\rowcolor{lightblue} 2-1& UAHN& rand.& 4& 1.0& None& 23.289& 0.4033& 0.5121& 0.3529& 0.5696& 0.4171& 1.7597 \\ 
				\rowcolor{lightblue} 2-2& CUAHN&rand.& 4& 10.0& None& -& 0.3747& 0.3529& 0.3249& 0.6017& 0.3884& 1.7813 \\ 
				\rowcolor{lightblue} 2-3& UAHN& rand.& 3& 0.5& None& 22.043& 0.4412& 0.5435& 0.3796& \textbf{0.5390}& 0.4288& 1.7786 \\	
				\rowcolor{lightblue} 2-4& UAHN\textbf{+}&rand.& 3& 10.0& None& -& 0.4053& 0.2965& \textbf{0.3145}& 0.5518& 0.4249& 1.7886 \\
				\rowcolor{lightblue} 2-5& CUAHN&rand.& 3& 10.0& None& -& 0.3496& 0.2930& 0.3195& 0.5954& 0.3950& 1.7869 \\	
				
				\midrule
				\rowcolor{lightblue} 3-1& UAHN&  Basic& 4& 30.0& None& -& 0.3628& 0.3827& 0.3779& 0.5823& 0.4290& 1.7732 \\
				\rowcolor{lightblue} 3-2& CUAHN& Basic& 4& 15.0& None& -& 0.3601& 0.3417& 0.3225& 0.5877& 0.4125& 1.7706 \\
				\rowcolor{lightblue} 3-3& UAHN&  Basic& 3& 30.0& None& -& 0.3606& 0.3614& 0.3738& 0.5866& 0.4329& 1.7821 \\
				\rowcolor{lightblue} 3-4& CUAHN& Basic& 3& 20.0& None& -& 0.3548& 0.3144& 0.3231& 0.5879& 0.4189& 1.7753 \\

				\midrule
				\rowcolor{lightgreen} 4-1& UAHN&  rand.& 3& 0.5& Ensem. (2\tnote{\dag} )& 26.075& 0.4664& 0.4497& 0.3494& 0.5725& 0.4156& 1.7731 \\
				\rowcolor{lightgreen} 4-2& CUAHN& rand.& 3& 5.0& Ensem. (2\tnote{\dag} )& -& 0.3493& 0.2846& 0.3206& 0.5865& 0.3937& 1.7880 \\
				\rowcolor{lightgreen} 4-3& UAHN&  rand.& 3& 0.5& Ensem. (3\tnote{\dag} )& 31.519& 0.4264&  0.3863& 0.3335& 0.5749& 0.4196& 1.7720 \\
				\rowcolor{lightgreen} 4-4& CUAHN& rand.& 3& 5.0& Ensem. (3\tnote{\dag} )& -& 0.3475& \textbf{0.2739}& 0.3200& 0.5826& 0.3985& 1.7903 \\
				
				\midrule
				\rowcolor{lightgreen} 5-1& UAHN& Basic& 3& 65.0& Ensem. (2\tnote{\dag} )& -& 0.3575& 0.3170& 0.3825& 0.6186& 0.4047& 1.8008 \\ 
				\rowcolor{lightgreen} 5-2& CUAHN& Basic& 3& 50.0& Ensem. (2\tnote{\dag} )& -& 0.3445& 0.3179& 0.3477& 0.6059& 0.4035& 1.7852 \\ 
				\rowcolor{lightgreen} 5-3& UAHN& Basic& 3& 50.0& Ensem. (3\tnote{\dag} )& -& 0.3662& 0.3126& 0.4199& 0.6148& 0.4033& 1.8052 \\ 
				\rowcolor{lightgreen} 5-4& CUAHN& Basic& 3& 30.0& Ensem. (3\tnote{\dag} )& -& 0.3544& 0.3072& 0.3353& 0.5992& 0.4042& 1.7811 \\ 
				
				\midrule
				\rowcolor{lightgreen} 6-1& UAHN& Basic\tnote{\ddag}& 4& 10.0& Drop. 5\% (16\tnote{\dag} )& 23.605& 0.3577& 0.3607& 0.3623& 0.5976& \textbf{0.3871}& 1.7903 \\
				\rowcolor{lightgreen} 6-2& CUAHN& Basic\tnote{\ddag}& 4& 10.0& Drop. 5\% (16\tnote{\dag} )& -& 0.3906& 0.3437& 0.3356& 0.6090& 0.3945& 1.7816 \\
				\rowcolor{lightgreen}
				& & & & 10.0& & & \textbf{0.3360}& 0.2976& 0.3761& 0.5989& 0.3970& 1.8003 \\
				\rowcolor{lightgreen}
				\multirow{-2}{*}{6-3}&\multirow{-2}{*}{UAHN}&\multirow{-2}{*}{Basic\tnote{\ddag}}&\multirow{-2}{*}{3}& 5.0& \multirow{-2}{*}{Drop. 5\% (16\tnote{\dag} )}& \multirow{-2}{*}{-}& 0.3371& 0.3139& 0.3392& 0.5837& 0.4066& 1.7905 \\
				
				\rowcolor{lightgreen}
				& & & & 10.0& & & 0.3436& 0.2915& 0.3417& 0.5959& 0.4067& 1.7854 \\
				\rowcolor{lightgreen}
				\multirow{-2}{*}{6-4}&\multirow{-2}{*}{CUAHN}&\multirow{-2}{*}{Basic\tnote{\ddag}}&\multirow{-2}{*}{3}& 5.0& \multirow{-2}{*}{Drop. 5\% (16\tnote{\dag} )}& \multirow{-2}{*}{-}& 0.3479& 0.3138& 0.3214& 0.5843& 0.4095& 1.7790 \\
				
				\rowcolor{lightgreen} 6-5& UAHN& Basic\tnote{\ddag}& 2& 5.0& Drop. 5\% (16\tnote{\dag} )& 19.419& 0.3533& 0.3015& 0.3359& 0.5935& 0.3964& 1.7881 \\
				\rowcolor{lightgreen} 6-6& CUAHN& Basic\tnote{\ddag}& 2& 5.0& Drop. 5\% (16\tnote{\dag} )& -& 0.3556& 0.3044& 0.3214& 0.5842& 0.4057& 1.7759 \\
				
				\rowcolor{lightgreen} 6-7& UAHN& Basic\tnote{\ddag}& 1& 5.0& Drop. 5\% (16\tnote{\dag} )& \textbf{17.455}& 0.5862& 0.3612& 0.3887& 0.6050& crash& 6.1602 \\
				\rowcolor{lightgreen} 6-8& CUAHN& Basic\tnote{\ddag}& 1& 5.0& Drop. 5\% (16\tnote{\dag} )& -& 0.4185& 0.3692& 0.3432& 0.6005& 0.4430& 8.8627 \\
				
				\bottomrule 
				
			\end{tabular}
			\begin{tablenotes}
				\footnotesize
				\item[$\star$] Average network inference time consumption, measured on a TX2 processor in Max-P ARM power mode. Networks with the same architectures were only measured once. For instance, the data of 3-2 is omitted since, theoretically, it should be the same as 2-1. 
				\item[\dag] The number of independent network models in an ensemble or the number of forward passes with dropout.
				\item[\ddag] Only to initialize the convolutional layers. The FC layers are randomly initialized and have dropout layers before them.
			\end{tablenotes}
		\end{threeparttable}
		}
	\end{table*}

	\subsection{Ablation Study} \label{Variants_CUAHN}
	
	In this ablation study, we aim to gain insights into how the components and setups of CUAHN-VIO contribute to VIO accuracy.
	The VIO variants are shown in Table \ref{CUAHN_variants_compare}. 
	The 2nd column shows the network acting as the vision frontend of a VIO variant. The 3rd column indicates the initialization method of a network.
	The last network block can be initialized randomly or by the Basic Model, as introduced in Subsection \ref{pred_uncertainty}. 
	The number of network blocks running in a VIO variant is shown in the 4th column. The 5th and 6th columns tell about the measurement covariance matrix $\bm{R}_{\text{meas.}}$
	and the estimation method of empirical uncertainty. 
	
	VIO variants are divided into six groups according to the shared setups.
	The VIO variants in Group 1 have no uncertainty estimation, which means that $\bm{R}_{\text{meas.}}$ stays constant for all network predictions. The shown value in the 5th column is the identical diagonal element of $\bm{R}_{\text{meas.}}$. For Group 2 to Group 6, uncertainty estimation is available. The 5th column shows the $k_{\text{var.}}$.
	Most EKF parameters, \textit{e.g.} $\bm{Q}$, stay fixed and are the same for all the VIO variants. $\bm{R}_{\text{meas.}}$ is the only manually-tuned parameter for different V
	For each VIO variant, we run it on Seq. 2 several times to find the $\bm{R}_{\text{meas.}}$ or $k_{\text{var.}}$ that yields the smallest ATE and use this value for other sequences. In practice, we found that the ATE is not very sensitive to $k_{\text{var.}}$.
	
	First, we look at the benefits of having predictive uncertainty. The VIO accuracy is improved substantially. This can be seen by comparing Group 1 (light yellow colored) with Group 2 and 3 (light blue colored).
	
	Second, we investigate
	whether it gives better results when empirical uncertainty is also estimated, by comparing Group 2 and 3 (light blue) with Group 4 to 6 (light green).
	Here, the differences are less pronounced. However, {most lowest ATEs are in light green groups, which do have empirical uncertainty.} Comparing deep ensembles (Group 4 and 5) with MC-Dropout (Group 6) does not lead to clear conclusions either. Given their similar accuracy, MC-Dropout is preferable due to its lower time consumption. {Another phenomenon we observed but is not shown in the table is that more than 16 times of sampling of MC-Dropout produces no noticeable improvement.}
	
	Third, we evaluate the effects of content-aware learning.
	CAHN of Group 1 is initialized by the Basic Model and further trained on the UZH-FPV training set in the same way as the networks in Subsection \ref{mask}.
	In other groups, CUAHN is trained with the aggregated dataset. Compare with UAHN which is trained with the Basic Dataset, CUAHN not only performs content-aware learning but also has seen more in-domain data, \textit{i.e.} the UZH-FPV training set.
	To see how much content-aware learning alone helps, we trained a master network on the aggregated dataset. 
	\eqref{self_loss} instead of \eqref{NLL_Lap} is the loss function thus it does not conduct content-aware learning.
	The student network 2-4 is trained by this master network on the aggregated dataset. The plus sign indicates that it has the bigger training set than other UAHNs.
	Comparing 2-4 and 2-5 that are trained on the same dataset, 2-5 is trained with the content-aware loss while 2-4 is not. 2-5 wins on four sequences out of six. But, in general, the differences are small. We conclude that the contribution of content-aware learning is small, the same as what is observed in Subsection \ref{mask}.
	
\begin{figure}[!hbpt]
		\centering
		\makebox{
			\includegraphics[scale=0.198]{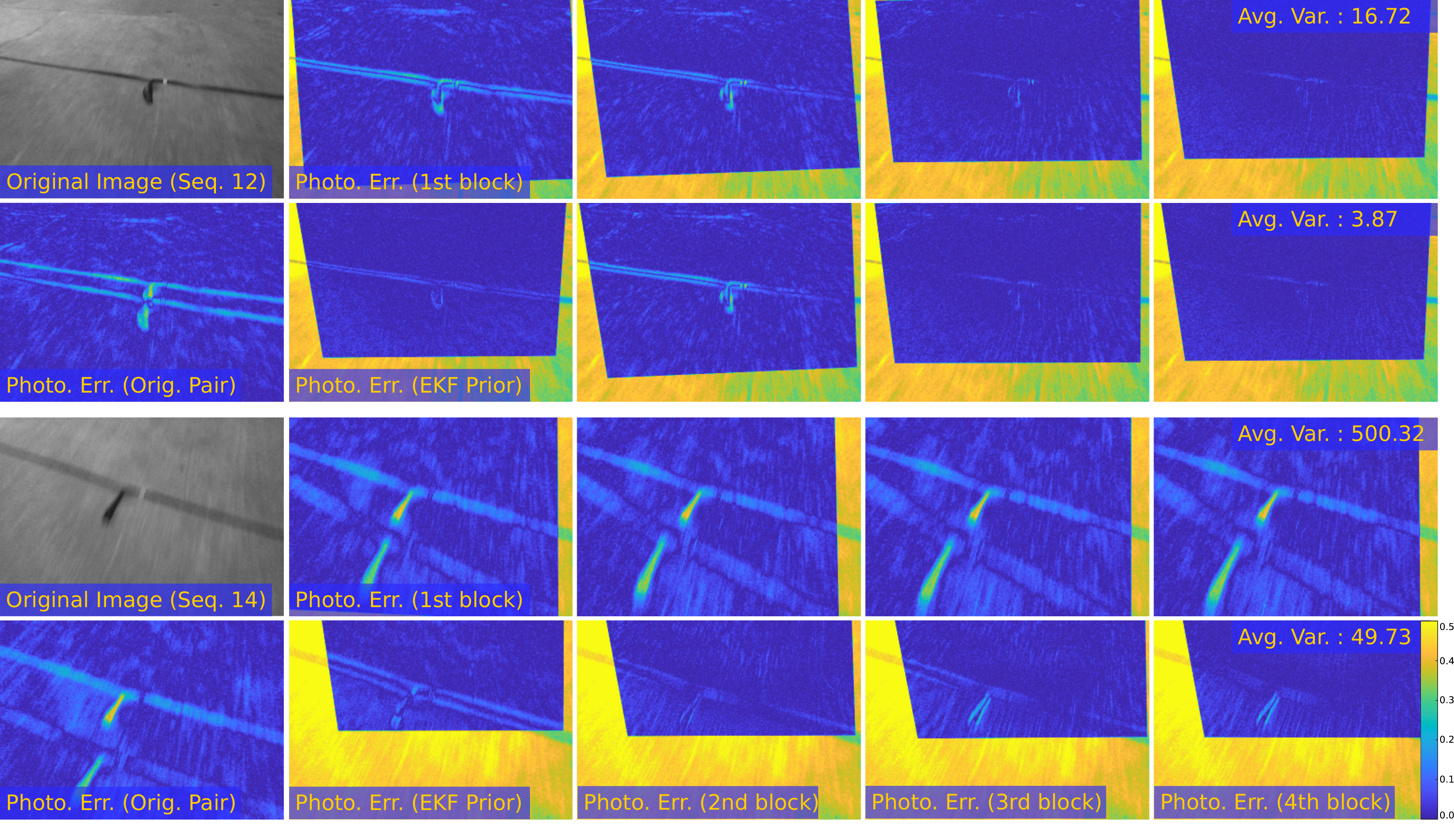}}
		\caption{The top two rows show an example image pair captured at a relatively slow speed (Seq. 12 of UZH-FPV). The bottom two rows show a high-speed example (Seq. 14). The two image pairs film the same scene. 
		The 1st column shows the original images and the original photometric error maps of the image pairs. 
		The 2nd column shows the photometric error maps of ($\widetilde{\mathcal{I}}_{t,1}$, $\mathcal{I}_{t-1}$) or ($\widetilde{\mathcal{I}}_{t,\text{prior}}$, $\mathcal{I}_{t-1}$). The 3rd to 5th columns show the photometric error maps of ($\widetilde{\mathcal{I}}_{t,i}$, $\mathcal{I}_{t-1}$), $i$ is the index of network block and ranges from 2 to 4.
		The performance of the network 6-2 in Table \ref{CUAHN_variants_compare} is shown in the 1st and 3rd rows, and network 6-4 in the 2nd and 4th rows.
		}
		\label{prior_slow_fast}
	\end{figure}

	Fourth, we assess the effects of exploiting \textit{a priori} homography for image pre-warping as shown in Fig. \ref{system_overview}. In Table \ref{CUAHN_variants_compare}, except for the VIO variants with four network blocks, \textit{a priori} homography is exploited for all other variants with less blocks.
	Running three blocks leads to comparable performance to four blocks with a small computational time gain ($\sim$1ms). 
	Reducing the number of blocks further leads to additional time gains. But using only one block results in worse VIO accuracy, as shown in Group 6. 
	Pre-warping with \textit{a priori} homography facilitates VIO accuracy especially in high speed, as illustrated in Fig. \ref{prior_slow_fast}.
	The 3rd row shows an example of high-speed flight. The 1st network block fails to predict the homography transformation well, which is not corrected by the subsequent blocks. Big estimated variances (top right of the rightmost image) indicate the network's low confidence in its prediction. The 4th row shows how the \textit{a priori} homography initializes the image pair. They are close to good alignment and further refined by the network blocks.
	
	Fifth, we study the influence of the initialization of the student network. Group 2 to Group 5 show no clear influence of this variable. This may mean that the trade-off between more accurate mean value prediction and better variance estimation is equitable and leads to similar VIO accuracy. We notice that the manually tuned parameter $k_{\text{var.}}$ is quite different between the initialization schemes. This tuning may be the partial cause of the similar accuracy. 
	The higher values of $k_{\text{var.}}$ for the ``Basic" initialization may compensate to a certain extent for the underestimation of uncertainty, although a simple scaling factor does not intrinsically improve the quality of uncertainty estimation. Since we think proper uncertainty estimation is one of the keys to good generalization and robustness, we have a light preference for random initialization. 
	
	\subsection{Onboard Deployment for Feedback Control}
	
	UAHN-VIO, 6-5 of Table \ref{CUAHN_variants_compare}, is deployed onboard an autonomous MAV to produce odometry information required by the close-loop feedback control. 
	Sparse non-planar objects are randomly laid on the planar floor of the flight arena. Images from this environment are not involved in network training.
	Thanks to the wide distribution of the Basic Dataset and the robustness toward non-planar content of the network as shown in Subsection \ref{mask}, theoretically, UAHN-VIO should work in any environment requiring no fine-tuning. 
	
	The MAV for flight experiments is a quadrotor equipped with a TX2 processor. An MYNT-EYE visual-inertial sensor is mounted 90-degree downward-facing. The image stream and IMU measurement stream are published at 30Hz and 200Hz respectively. UAHN-VIO subscribes to sensor data and publishes the estimated attitude, velocity, and position once it has processed the latest image. So the odometry information has the same frequency as the image stream. We did not compensate for the processing latency\footnotemark[\value{footnote}] because it is low and stable, as introduced later. 
	The flight controller is a basic proportional-integral-derivative (PID)-based position and velocity controller. It generates thrust and attitude control commands that are sent to Betaflight4 for low-level control.
	
	We tested three kinds of flights, hover\footnotemark[\value{footnote}], tracking a circle trajectory\footnotemark[\value{footnote}], and shuttle flight between two waypoints. During autonomous flights, the sensor data was recorded for offline replay. During the two-waypoint shuttle flight, the controller was badly tuned on purpose to induce larger motions, resulting in a variety of captured images.
	The velocity and trajectory plots of the two-waypoint shuttle flight are shown in Fig. \ref{mynt_shuttle}. The link to the flight video is in \textit{Supplementary Materials}.
	
	\subsection{Time Efficiency and Processing Latency}
	
	\begin{table*}[!htbp]
		\centering
		\setlength{\tabcolsep}{0.5mm}{
			\caption{Time Consumption Indicators Measured on TX2 Processor by Running Seq. 2 of UZH-FPV. \textbf{Bold}
				Represents the Best. \underline{Underline} marks the Frame Rates and ATEs valid for Accuracy Evaluation.}
			\label{time_consume_procedures}
			\centering
			\begin{threeparttable}
				\begin{tabular}
					{m{40pt}<{\centering}m{35pt}<{\centering}m{40pt}<{\centering}m{41pt}<{\centering}m{38pt}<{\centering}m{39pt}<{\centering}m{35pt}<{\centering}m{36pt}<{\centering}m{36pt}<{\centering}m{48pt}<{\centering}m{50pt}<{\centering}m{36pt}<{\centering}}
					\toprule 
					VIO& Image Resolution (pixels)& Num. of Pts / Network Blocks& Histogram Equalization (HE) & Visual Processing Time (ms)& IMU Propagation Time (ms)& EKF Updating Time (ms)& Total Time Mean (ms) $\downarrow$& Total Time Variance (ms$^2$) $\downarrow$& Ratio of Long Processing Time (\%) $\downarrow$& Avg. Processed Frame Rate (fps) $\uparrow$& RMSE (m) of ATE $\downarrow$\\
					
					\midrule
					MSCKF& 640$\times$480& 300& \checkmark&38.37& 19.38& 8.43& 66.19& 3.95e4& 66.30& 18.06& 0.3142 \\
					
					MSCKF& 320$\times$240& 180& \checkmark& 24.47& 2.29& 6.04& 32.80& 471.61& 25.94& 23.78& 0.4058 \\
					
					MSCKF& 640$\times$480& 100& \checkmark&26.14& 2.47& 3.94& 32.57& 549.43& 29.88& 23.14& 0.3386 \\
					
					MSCKF& 640$\times$480& 100& \usym{2613}& 22.85& 2.09& 3.80& 28.74& 460.15& 24.93& 24.24& 0.3178 \\
					
					MSCKF& 640$\times$480& 10& \checkmark& 15.81& 1.66& 1.28& 18.75& 233.38& 12.21& \underline{26.23}& \underline{0.4112} \\
					
					MSCKF& 320$\times$240& 10& \checkmark& \textbf{11.57}& 1.78& 1.48& \textbf{14.83}& 157.05& 7.01& \underline{26.23}& \underline{0.6000} \\
					\midrule
					UAHN-VIO& 320$\times$224& 2& \usym{2613}& 24.00& 1.48& 0.12& 25.61& 3.32& \textbf{0.44}& \underline{26.11}& \underline{0.3544} \\
					UAHN-VIO& 320$\times$224& 3& \usym{2613}& 27.30& \textbf{1.46}& \textbf{0.12}& 28.89& \textbf{2.66}& 0.78& \underline{26.11}& \underline{\textbf{0.3380}} \\
					
					\bottomrule
					
				\end{tabular}
				\footnotesize
			\end{threeparttable}
		}
	\end{table*} 
	
	We have shown the network inference time consumption in Table \ref{CUAHN_variants_compare}. In the following, we further discuss the detailed time consumption and processing latency of the whole VIO system. 
	The 1st row of Table \ref{time_consume_procedures} shows the time-consumption-related indicators. We log the time consumption of the three main computing procedures (visual processing, IMU propagation, and EKF updating) of each frame. The mean and variance of the total time consumption are calculated. Besides, we compare the total time cost of processing each frame with the standard time interval of the 30Hz video (33.3ms). The 3rd column from the right shows the percentage of frames that take more than 33.3ms in all frames.  
	In the implementation of both MSCKF and UAHN-VIO, only when a frame has been processed, the filter state at this timestamp is recorded and used to calculate ATE. 
	The Average Processed Frame Rate (2nd column from the right) equals to the number of processed frames divided by the video duration in seconds. 
	This is a metric for how many frames are skipped. The cause of skipping a frame is the limited fixed size of the image buffer. In the implementation,
	if the VIO processing is too slow and more than five images are waiting for processing in the buffer, the oldest one will be discarded to make room for the new image. 
	
	The MSCKF \cite{li2013high} in Table \ref{time_consume_procedures} is implemented by \cite{geneva2020openvins}. We choose it instead of other SOTA approaches because our C++ implementation is based on the open-sourced code of \cite{geneva2020openvins}. Thus the comparison is as fair as possible due to the similarities in code. Besides, the efficiency of MSCKF is also advantageous as a filter-based approach. 
	We change the image resolution, the number of processed feature points, and histogram equalization of the MSCKF because all of them observably affect the time consumption.
	The data in Table \ref{time_consume_procedures} is measured on a TX2 processor in the power mode that the VIO approach runs faster. MSCKF computes everything with the CPU. It runs faster in the Max-N power mode. The network of UAHN-VIO runs on GPU and is faster in Max-P ARM mode. According to our observation, it applies to CUDA-accelerated DNNs implemented in PyTorch and LibTorch running on TX2 processors.
	
	The MSCKF of the 2nd row of Table \ref{time_consume_procedures} has the same settings as the one in Table \ref{SOTA_VIO}. When running on a TX2 processor, the average time cost of processing a frame is around two times the standard time interval. We reduce the number of feature points and downscale the images to lower the average time to just below 33.3ms, as shown in the 3rd and 4th rows. But the variance of the total time is still very big. 25\% to 30\% frames require a longer time than 33.3ms to process. And there are still frames skipped. The 4th and 5th rows tell us that the better robustness toward motion blur (to be introduced later) brought by histogram equalization comes at an expense of $\sim$3.3ms extra processing time.
	
	The number of feature points is further reduced to only ten as shown in the 6th and 7th rows to minimize the time cost.
	Only the bottom four rows of Table \ref{time_consume_procedures} manage to process all the frames. The subtle difference between 26.23 and 26.11 is caused by the different stop time of the two VIO approaches.
	The frame rates are less than 30 because irregular frame drops exist in the original 30-fps video of UZH-FPV.
	Comparing ATEs of trajectories at very different frequencies is not informative. So the ATEs in the top four rows are out of the discussion.
	Comparing the ATEs of UAHN-VIO in Table \ref{time_consume_procedures} and Table \ref{SOTA_VIO}, they are almost the same on different machines. 
	The accuracy of MSCKF processing ten points is only slightly worse than when processing 300 points as shown in Table \ref{SOTA_VIO}. 
	
	It is clear that for both approaches in Table \ref{time_consume_procedures}, visual processing takes most of the time. For MSCKF, the visual processing time is significantly affected by the number of feature points and image resolution. The variance of total time decreases with the mean value, but it is still very big compared with the ones of UAHN-VIO. Even if the number of points is only ten, there are still around 10\% of frames whose processing cannot be finished before the next frame comes. In comparison, UAHN-VIO's variance of total time cost is very small, which indicates that the processing time is almost constant for every frame.
	The very few (less than 1\%) frames that take longer processing time for UAHN-VIO are the first several of the video. The cause of it is likely to be library related, \textit{i.e.}, the warm-up phase of the network object of LibTorch. 
	Besides the stable network inference time, UAHN-VIO has low and constant state propagation and updating time cost because of its very simple filter design. Most computation is the network inference on the GPU. As a result, the CPU usage of UAHN-VIO is very low.
	
	\begin{figure}[!hbpt]
		\centering
		\makebox{
			\includegraphics[scale=0.58]{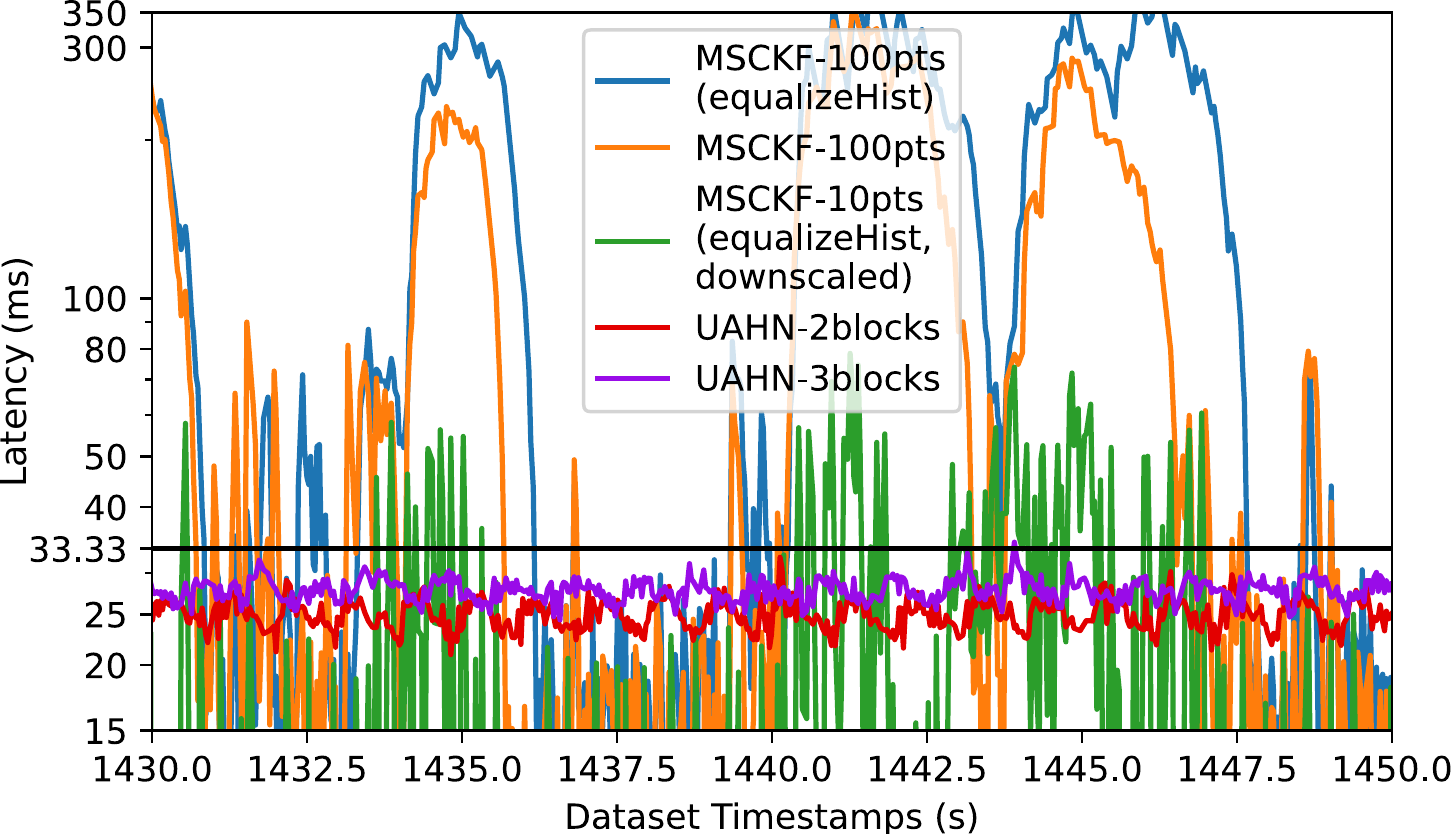}}
		\caption{Processing latency within 20 seconds of Seq. 2, UZH-FPV dataset. 
		}
		\label{latency}
	\end{figure}
	
	Fig. \ref{latency} shows the processing latency measured on a TX2 processor. It is the time gap between capturing a new image and updating the filter states according to the image.
	The latency of UAHN-VIO is very stable thanks to the image-content-independent network inference time cost and the simple filter design. The big variation in the latency of MSCKF corresponds to the big variance of time consumption shown in Table \ref{time_consume_procedures}. MSCKF using original images (yellow curve) has smaller latency than its peer that conducts histogram equalization (blue curve). When the number of points is reduced to only ten and the images are downscaled to half resolution (green curve), latency significantly decreases but is still noisy and often beyond 33.3 ms.
	
	The above discussion about the processing time consumption of MSCKF only applies to the current CPU implementation. 
	There are GPU-based implementations for handcrafted feature points such as \cite{heymann2007sift}, and learning-based feature points \cite{laguna2022key, detone2018superpoint, sarlin2020superglue}. 
	VIO approaches based on feature points adopting such techniques can achieve lower and scene-independent stable latency in their vision frontends. 
	But the complicated backends that utilize the pixel trajectories of the vanished points \cite{li2013high}, BA \cite{qin2017vins}, or iterative EKF \cite{bloesch2017iterated} still require serial computing and the required CPU resources can be considerable and scene-dependent. 
	As far as we know, most traditional VIO approaches only have CPU implementations.
	So before their GPU versions are widely recognized, CUAHN-VIO has an advantage in processing latency.
	Besides, it requires small CPU resources and thus allows the deployment of computationally heavy iterative planning and control approaches that run better on CPUs. 
	
	\subsection{Robustness toward High-Speed Flight} \label{blur_robust}
	
\begin{figure}[!hbpt]
		\centering
		\makebox{
			\includegraphics[scale=0.45]{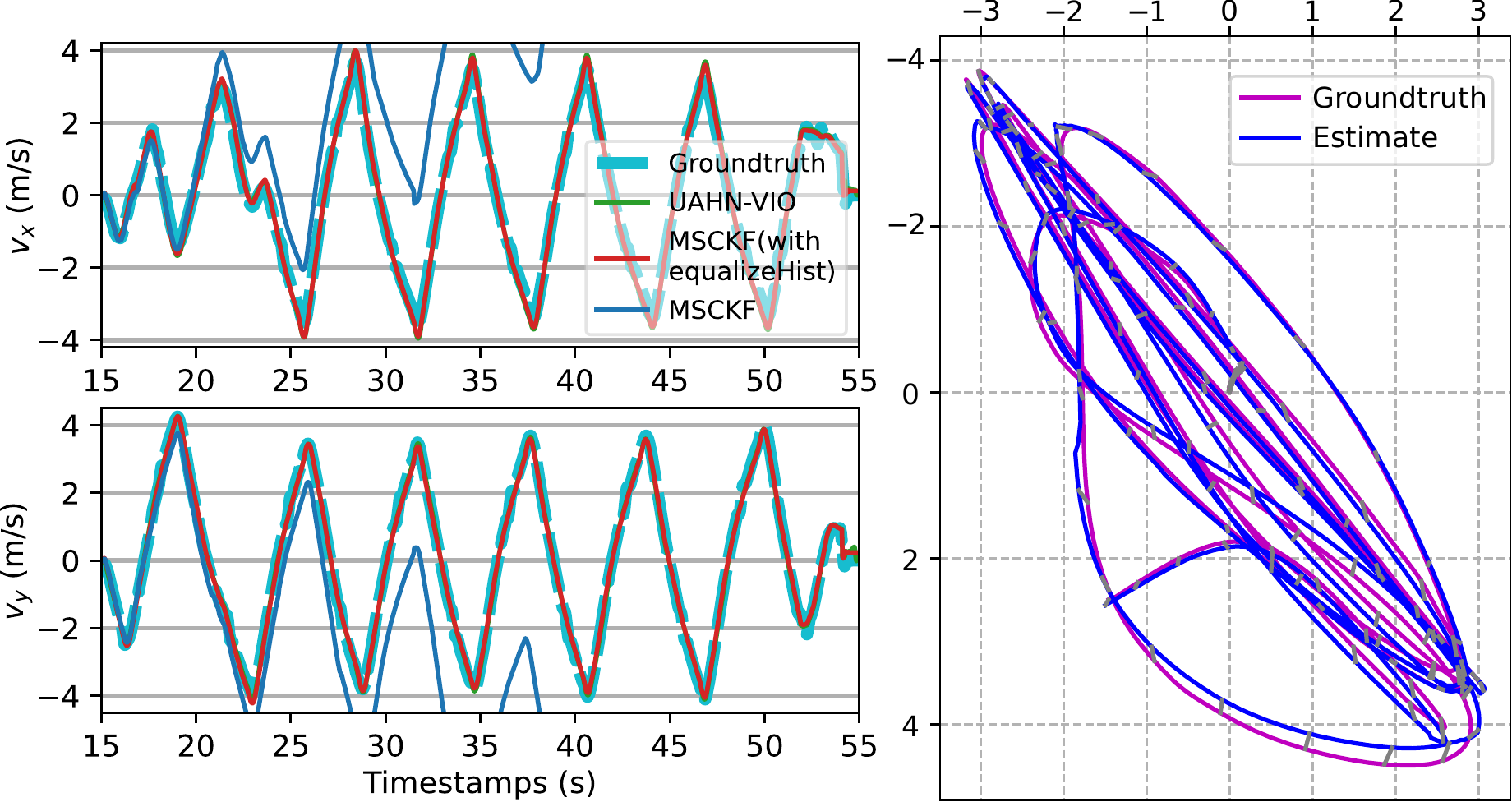}}
		\caption{UAHN-VIO (6-5 in Table \ref{CUAHN_variants_compare}) and MSCKF (100 feature points) with and without histogram equalization are evaluated on the two-waypoint shuttle flight sequence, running on a TX2 processor. The left subplot shows the velocity expressed in the body frame. The right subplot is the trajectory evaluation of UAHN-VIO plotted by \cite{zhang2018tutorial}. The groundtruth was recorded by an Optitrack motion capture system. The average and maximum speeds during the flight are respectively 2.87 m/s and 5.41 m/s. The distance to the ground ($z$-axis) is stabilized at one meter.
		}
		\label{mynt_shuttle}
	\end{figure}

	A bad effect of high-speed flight on VIO is the huge optical flow in the image plane, especially when the distance to the ground is small. Due to the fixed sample interval and non-neglectable exposure duration of a frame-based camera, visual disparities between consecutive images and motion blur correspondingly grow with optical flow. Regarding big visual disparities, in Fig \ref{prior_slow_fast}, we show a failure case that is solved by the \textit{a priori} homography. Confronting motion blur, in the following, we demonstrate the advantage of using a network as the vision frontend over processing handcrafted feature points.
	Same as before, we compare UAHN-VIO and MSCKF.
	A big percentage of images captured during the two-waypoint shuttle flight (Fig. \ref{mynt_shuttle}) have significant motion blur thus this sequence is used to evaluate the robustness toward blur. The quadrotor MAV maximized its tilt angle to speed up and down. When the speed reached a peak, the MAV rotated to slow down. In this case, the optical flow was caused together by the fastest translation and rotation thus it achieved a peak. 
	
	\begin{figure}[!hbpt]
		\centering
		\makebox{
			\includegraphics[scale=0.59]{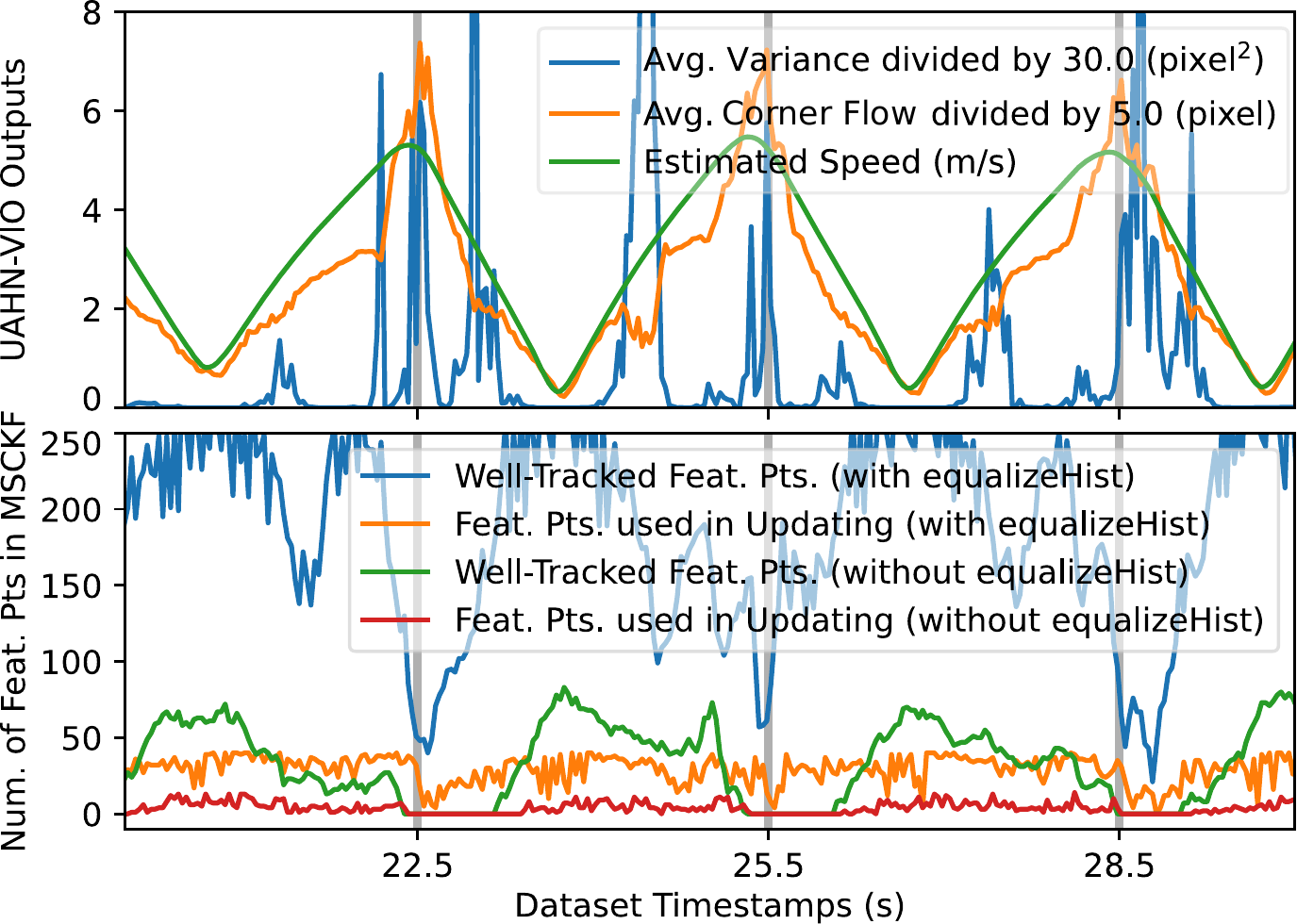}}
		\caption{The outputs of UAHN-VIO (6-5 of Table \ref{CUAHN_variants_compare}) and MSCKF (2nd row of Table \ref{time_consume_procedures}) processing the two-waypoint shuttle flight sequence. 
		The average of the network-estimated variances of the eight elements of the corner flow and the average of the absolute values of the eight elements of the corner flow are downscaled for better illustration. The well-tracked points are the ones that fulfil the epipolar constraint calculated within a RANSAC scheme.
		}
		\label{fast_num_pts}
	\end{figure}
	
\begin{figure}[!hbpt]
		\centering
		\makebox{
			\includegraphics[scale=0.2]{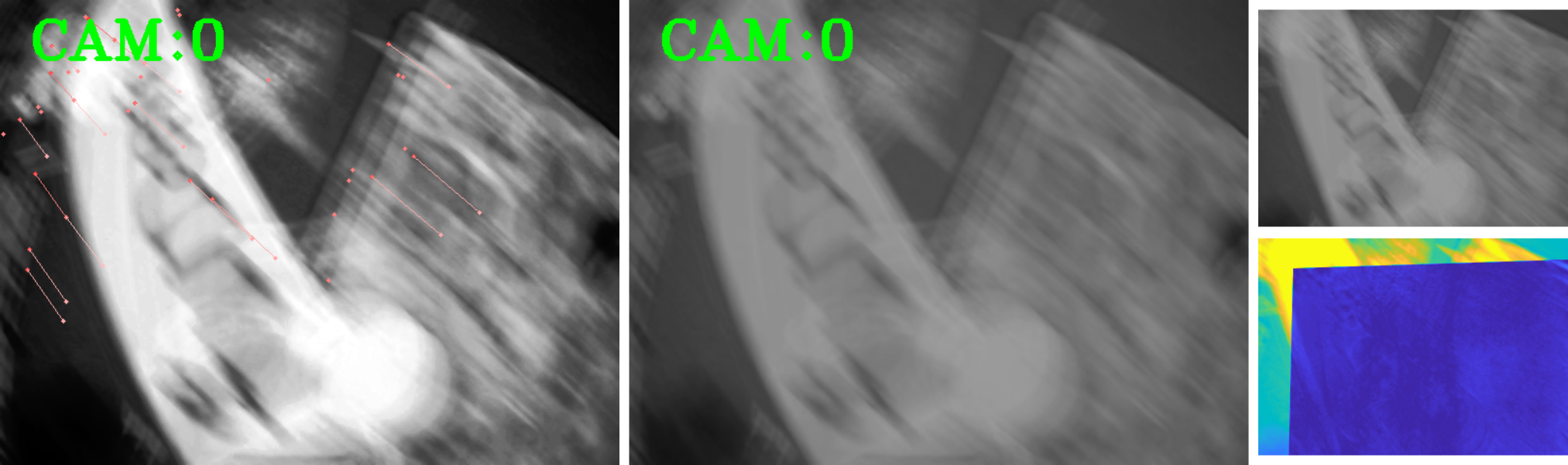}}
		\caption{An example of the highly blurry images captured during the two-waypoint shuttle flight.
		Feature point tracking results of MSCKF with and without histogram equalization are shown on the left and middle respectively. These two images are from the visualization module of \cite{geneva2020openvins}. In the left image, there are a few point tracking trajectories visualized by small points and thin line segments in red color.
		There is no point tracking trajectory in the middle image. The top right is the undistorted and resized image that is the input of the network. The bottom right is the photometric error map corresponding to the prediction of network 6-5 in Table \ref{CUAHN_variants_compare}. Dark blue indicates small photometric error.
		The average of the network-estimated variances of the eight elements of the corner flow is 31.12 (pixel$^\text{2}$). Images in this figure are shown in the actual resolutions when being fed into the VIO approaches. The network uses the smaller resolution.
		}
		\label{mynt_blur}
	\end{figure}

	A well-known problem of handcrafted visual feature points is that detection and tracking become more difficult in the presence of growing motion blur. 
	The bottom subplot of Fig. \ref{fast_num_pts} shows the sharp declines in the number of points when optical flow was around its peaks. 
	With histogram equalization, the number of points drops to less than 20\% of before.
	Without histogram equalization, the number drops to and stays at zero until the speed is slow enough. The lack of visual updating causes the MSCKF to drift as shown in the left subplot of Fig. \ref{mynt_shuttle}.
	Fig. \ref{mynt_blur} shows an image captured when the optical flow is close to a peak. 
	It is too blurry for FAST feature point \cite{rosten2006machine} that relies on local gradients. Histogram equalization increases the image gradients and produces several points without lowering the threshold for feature detection and tracking. But it also induces noise.
	Most point trajectories only have two frames and very few have three, which indicates that it is hard to keep tracking the already hard-to-detect points.
	In contrast, despite the severe reduction of local gradients, there are remaining gradients at bigger scales that can be captured by the network. 
	As shown by the photometric error map (bottom right of Fig. \ref{mynt_blur}), the network is able to retrieve a reasonable homography transformation that aligns the images well.
	
	A clear phenomenon shown by the top subplot of Fig. \ref{fast_num_pts} is that the network is more likely to have big uncertainty estimation at high speed. On most occasions when the network-estimated variance grows, the number of feature points dramatically declines or is already low, which is an indicator of emerging motion blur. 
	Motion blur can be treated as noise in the input of the network and thus it outputs big predictive uncertainty.
	Fig. \ref{prior_slow_fast} shows an example that for similar image content with different amounts of blur, based on the reasonable ranges speculated from the photometric error map, the variance is overestimated when the blur is more. 
	Overestimated variances make the relatively accurate mean predictions less trusted in the measurement updates of EKF and thus cause suboptimal results.
	More examples of uncertainty estimation for blurry images and the positive correlation between speed and estimated uncertainty are available in \textit{Supplementary Document} for interested readers.
	
	To summarize, VIO approaches that utilize feature points would not necessarily drift because of the lack of points caused by motion blur.
	Histogram equalization as image pre-processing can significantly increase the number of useful points.
	Besides, the well-designed VIO backends compensate for the effect of fewer points to some extent.
	We did not use datasets with long periods of ongoing motion blur in this article, but expect that these would be more problematic for feature-based approaches.
	The network suffers from the overestimated uncertainty caused by motion blur. But we have not observed that the accuracy of mean prediction is noticeably affected.
	Also because of the higher accuracy of UAHN-VIO than other approaches in most tests in this article, we believe that the proposed network has advanced robustness toward motion blur.
	
	\subsection{Potential Improvements}
	
	About reaching the end of this article, we discuss ideas that potentially improve the performance of CUAHN-VIO but have not been implemented yet.
	CUAHN-VIO has a toy-like EKF that only updates the current filter state. Involving states at more time steps into a sliding window filter and performing BA may help achieve smoother trajectories. 
	Keyframe achieved big success in geometry-based VIO. It can also be applied to CUAHN-VIO following the similar scheme proposed in \cite{zhong2020efficient}. 
	To handle a planar surface that is not orthogonal to the gravity vector, for example, a slope, the EKF can be extended according to \cite{zhong2020direct} and \cite{zhong2020efficient}.
	
	In training, the four network blocks are trained to handle the whole homography transformation. 
	When the \textit{a priori} corner flow propagated by the EKF is utilized for pre-warping, the distribution of the network input can be different from the training set. It can be helpful to fine-tune the network when running the whole VIO on a video.
	We do not implement this idea mainly due to the concern of overfitting to the scene and motion pattern of the fine-tuning videos.
	
	\section{Conclusions}
	
	In this article, we propose CUAHN-VIO. Its vision frontend is a homography transformation network with uncertainty awareness and the backend is a simple EKF. Evaluations show its comparable accuracy to SOTA traditional VIO approaches and its advantages in processing latency. The robustness toward motion blur, a trait of learning-based approaches, is observed again in this article. The synthetic big-scale training set is proven to enable a homography network to generalize well to the real world.
	Comparative studies show that, in our context, content-aware learning helps the accuracy to a small extent while uncertainty estimation from the network contributes significantly. 
	Most importantly, different from pursuing better performance through deeper networks and complicated loss functions, this work points out that, without requiring ground truth, a small-size network with a practical training scheme for uncertainty estimation can also stand out.
	
	\bibliographystyle{ieeetr}
	\bibliography{reference.bib}
	
\end{document}